%% file: main.tex
\documentclass[runningheads]{llncs}

\usepackage{eccv}

\makeatletter
\def\@fnsymbol#1{\ensuremath{\ifcase#1\or \dagger\or \ddagger\or
   \mathsection\or \mathparagraph\or \|\or **\or \dagger\dagger
   \or \ddagger\ddagger \else\@ctrerr\fi}}
\makeatother

\input{cmd.tex}

\def\ourName{TORS\xspace}

\usepackage{eccvabbrv}

\usepackage{graphicx}
\usepackage{booktabs}

\usepackage[accsupp]{axessibility}  

\usepackage{hyperref}

\usepackage{orcidlink}

\begin{document}

\title{Analyzing and Improving Training-Free Fast Sampling of Text-to-Image Diffusion Models} 
\titlerunning{Analyzing and Improving Fast Sampling of Text-to-Image Diffusion Models}

\author{Zhenyu Zhou\inst{1,2} \and Defang Chen\inst{3,4}\thanks{Corresponding author} \and Siwei Lyu\inst{4} \and Chun Chen\inst{1,2} \and Can Wang\inst{1,2}}

\authorrunning{Zhou et al.}

\institute{
$^1$State Key Laboratory of Blockchain and Data Security, Zhejiang University \\
$^2$Hangzhou High-Tech Zone (Binjiang) Institute of Blockchain and Data Security
$^3$University of California, Berkeley \quad $^4$State University of New York at Buffalo
}
\maketitle
\input{sec/abs}
\input{sec/intro}
\input{sec/preliminary}

\input{sec/method}
\input{sec/exp}

\input{sec/conclusion}

\section*{Acknowledgements}
Zhenyu Zhou and Can Wang are supported by the National Natural Science Foundation of China (No. 62476244), the Starry Night Science Fund of Zhejiang University Shanghai Institute for Advanced Study, China (Grant No: SN-ZJU- SIAS-001) and the advanced computing resources provided by the Supercomputing Center of Hangzhou City University.

%
%
\bibliographystyle{splncs04}
\bibliography{main}

\input{sec/X_suppl}

\end{document}

%% file: cmd.tex
\usepackage{graphicx}
\usepackage{wrapfig}
\usepackage{amsmath}
\usepackage{amssymb}
\usepackage{booktabs}
\usepackage{subcaption}
\usepackage{mathtools}
\usepackage{pifont}
\usepackage{cancel}
\usepackage{algorithm}
\usepackage{algorithmic}
\usepackage{multirow}
\usepackage{colortbl}
\usepackage{bbding}
\usepackage{utfsym}
\usepackage{array}
\usepackage{makecell}

\def\bfc{\mathbf{c}}

\def\bfr{\mathbf{r}}

\def\bfu{\mathbf{u}}

\def\bfw{\mathbf{w}}
\def\bfx{\mathbf{x}}

\def\bfT{\mathbf{T}}
\def\bfN{\mathbf{N}}
\def\bfB{\mathbf{B}}

\def\bfI{\mathbf{I}}

\def\rmd{\mathrm{d}}
\def\rmdx{\mathrm{d}\bfx}

\def\bbE{\mathbb{E}}
\def\bbR{\mathbb{R}}

\def\eqref#1{Eq.~(\ref{#1})}


%% file: sec/abs.tex
\begin{abstract}
Text-to-image diffusion models have achieved unprecedented success but still struggle to produce high-quality results under limited sampling budgets. Existing training-free sampling acceleration methods are typically developed independently, leaving the overall performance and compatibility among these methods unexplored. In this paper, we bridge this gap by systematically elucidating the design space, and our comprehensive experiments identify the sampling time schedule as the most pivotal factor. Inspired by the geometric properties of diffusion models revealed through the Frenet-Serret formulas, we propose \textit{cons\textbf{t}ant t\textbf{o}tal \textbf{r}otation \textbf{s}chedule} (\ourName), a scheduling strategy that ensures uniform geometric variation along the sampling trajectory. \ourName outperforms previous training-free acceleration methods and produces high-quality images with 10 sampling steps on Flux.1-Dev and Stable Diffusion 3.5. 
Extensive experiments underscore the adaptability of our method to unseen models, hyperparameters, and downstream applications. Code is available at \url{https://github.com/zju-pi/TORS}.


\end{abstract}

%% file: sec/intro.tex
\section{Introduction}
\label{sec:intro}


Diffusion-based generative models~\cite{sohl2015deep,song2019ncsn,ho2020ddpm} synthesize data by progressively denoising noisy samples, which can be modeled using ordinary differential equations (ODEs)~\cite{song2021sde}. Despite their impressive generative capabilities, diffusion models typically require hundreds of sampling steps, making sampling acceleration a long-standing research topic~\cite{ho2020ddpm,song2021sde,rombach2022ldm}. Existing acceleration methods can be categorized into training-based and training-free methods. Training-based methods fine-tune the original model to distill multi-step sampling into fewer steps~\cite{salimans2022progressive,meng2023distillation,song2023consistency,yin2023one,zhou2024simple}. However, such methods become increasingly resource-intensive and costly as the model sizes grow exponentially, with state-of-the-art text-to-image diffusion models reaching billions of parameters (\eg, 8B Stable Diffusion 3.5~\cite{esser2024scaling}, 12B FLUX~\cite{flux}, 20B Qwen-Image~\cite{wu2025qwenimagetechnicalreport}, 24B Playground v3~\cite{liu2024playground}). To improve deployment efficiency without additional training, recent work has shifted toward training-free approaches, such as fast ODE solvers~\cite{song2021ddim,karras2022edm,lu2022dpm,zhang2023deis,zhao2023unipc,zhou2023fast}, optimized time schedules~\cite{nichol2021improved,sabour2024align,chen2024trajectory,xue2024accelerating}, and feature caching~\cite{ma2024deepcache,wimbauer2024cache,li2023faster,liu2025reusing}. 

However, existing training-free methods have largely been developed in isolation, and no systematic study, particularly for state-of-the-art text-to-image diffusion models, has examined their design components or compared their relative contributions to overall performance. In this paper, we bridge this gap by elucidating the design space of ODE solvers, time schedules, and feature caching methods. Through comprehensive experiments on five key components of training-free acceleration methods, we identify the time schedule as the most pivotal factor affecting model performance, while the default uniform time schedule is suboptimal, leading to slow convergence in image structure. Based on this observation, we introduce an improved time scheduling strategy that leverages the geometric properties of sampling trajectories, specifically curvature and torsion under the Frenet-Serret formulas. Our method enforces a constant total geometric variation along the sampling trajectory, yielding high-quality images with consistent structure that closely match converged multi-step results with just a few steps. 
Moreover, we verify the adaptability of our method by transferring it to unseen LoRA-fine-tuned variants, architectures, hyperparameters, and downstream applications such as image editing.
Finally, we conduct a comprehensive compatibility evaluation across various training-free acceleration methods, demonstrating performance improvements and strong robustness of our method. In summary, our contributions are as follows:

\begin{figure*}[t]
    \centering
    \includegraphics[width=\textwidth]{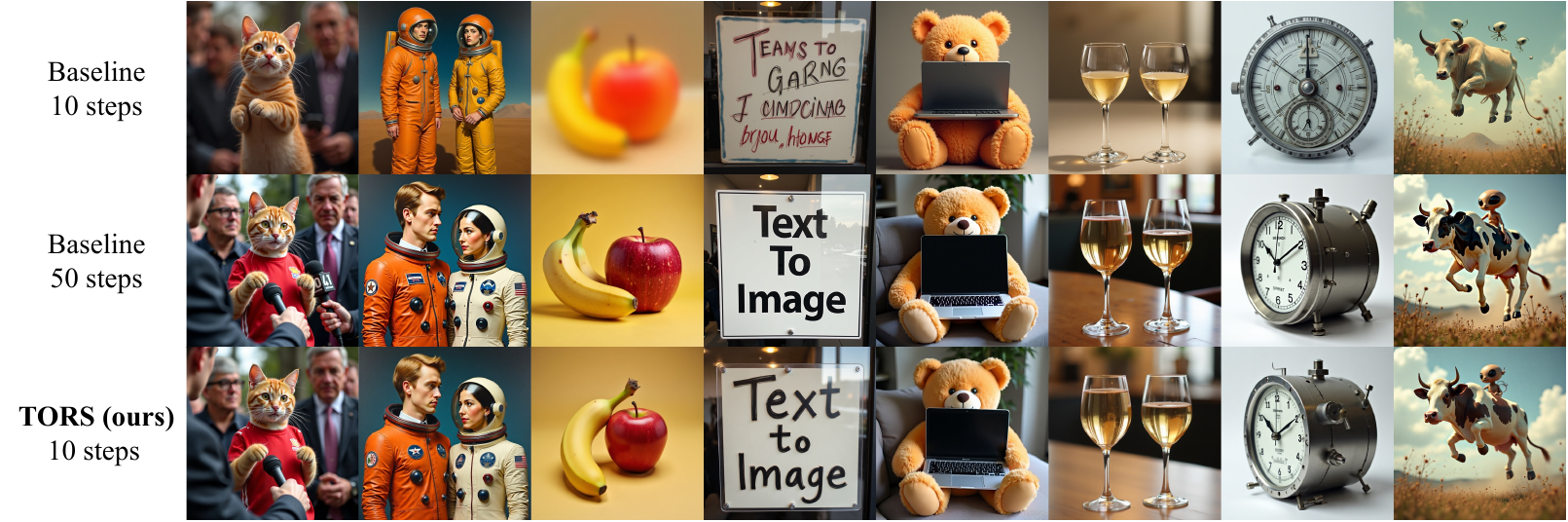}
    \caption{Text-to-image generation using Flux.1-Dev~\cite{flux}. Our proposed \ourName attains image quality comparable to the 50-step baseline using only 10 steps.
    }
    \label{fig:teaser}
\end{figure*}

\begin{itemize}
    \item We elucidate the design space of mainstream training-free acceleration methods, quantify the impact of each component on sampling acceleration, and pinpoint the outer time schedule as the dominant factor.
    \item We propose an advanced scheduling strategy, \ourName, based on the geometric regularity of flow-based text-to-image sampling trajectories, which maintains constant total variation for stable and efficient generation.
    \item Extensive evaluation demonstrating that \ourName consistently yields significant performance gains while exhibiting architecture-agnostic adaptability, hyperparameter robustness, and seamless task transferability.
\end{itemize}

%% file: sec/preliminary.tex
\section{Background}


Given a data sample $\bfx_0 \in \bbR^d$ from the target data distribution $p_0$, the forward process in diffusion models gradually adds Gaussian noise to the sample, following a stochastic differential equation (SDE)~\cite{song2021sde}: 
\begin{equation}
    \rmd \bfx_t = f(t)\bfx_t\rmd t + g(t) \rmd \bfw_t, \quad t \in \left[0, T\right],
\end{equation}
where $f(t): \bbR \rightarrow \bbR$, and $g(t): \bbR \rightarrow \bbR$ are drift and diffusion coefficients, and $\bfw_t \in \bbR^d$ is the Wiener process~\cite{oksendal2013stochastic}. 
The backward process reconstructs the data through a reverse-time SDE, $\rmdx_t = [f(t)\bfx_t- g^2(t)\nabla_{\bfx_t} \log p_{t}(\bfx_t)]\rmd t + g(t) \rmd \bar{\bfw}_t$, which shares the same marginal distributions $\{p_t\}_{t=0}^T$ with the forward process~\cite{song2021sde}. This reverse-time SDE has a \textit{probability flow} ordinary differential equation (PF-ODE) counterpart~\cite{song2021sde,chen2024trajectory}, $\rmdx_t = [f(t)\bfx_t- \frac{1}{2}g^2(t)\nabla_{\bfx_t} \log p_{t}(\bfx_t)]\rmd t$. 
The analytically intractable $\nabla_{\bfx_t} \log p_{t}(\bfx_t)$ is known as the \textit{score function}~\cite{hyvarinen2005estimation}, which is the target to be predicted by diffusion models. Depending on the choice of $f(t)$ and $g(t)$, two specific forms of linear SDEs, namely, the variance-preserving (VP) SDE~\cite{ho2020ddpm} and the variance-exploding (VE) SDE~\cite{song2019ncsn} are widely used. 

In contrast, flow-based models~\cite{liu2022flow,lipman2022flow} aim to use a time-dependent \textit{velocity filed} $\bfu_t(\cdot): \bbR^d \rightarrow \bbR^d$ to generate a probability path between a data distribution $p_0$ and a noise distribution $p_T=\mathcal{N}(\bf0,\bfI)$ by
\begin{equation}
    \label{eq:flow_mathcing}
    \rmd \bfx_t = \bfu_t(\bfx_t)\rmd t, \quad t \in \left[0, T\right].
\end{equation}
Without loss of generality, we set $T=1$ thereafter. To eliminate the intractable velocity field, \textit{conditional velocity field} $\bfu_t(\cdot|\bfx_0)$ is used to define the generative forward process. We have $\bfu_t(\bfx_t|\bfx_0) = \frac{\dot{\sigma_t}}{\sigma_t}\bfx_t + \alpha_t \left( \frac{\dot{\alpha_t}}{\alpha_t} - \frac{\dot{\sigma_t}}{\sigma_t} \right) \bfx_0$ given $\bfx_t = \alpha_t \bfx_0 + \sigma_t \bfx_1$.
Instead of regressing the intractable velocity field with the objective $\bbE_{t,p_t(\bfx_t)}\lVert \bfu_\theta(\bfx_t) - \bfu_t(\bfx_t) \rVert_2^2$, the model is trained by conditional flow matching objective 
$\bbE_{t,p_t(\bfx_t|\bfx_0),p_0(\bfx_0)}\lVert \bfu_\theta(\bfx_t) - \bfu_t(\bfx_t|\bfx_0) \rVert_2^2$. The backward process is simply reversing \Cref{eq:flow_mathcing} in time and replacing the velocity by a well-trained model $\bfu_\theta$. For text-to-image generation, an additional text prompt $\bfc$ is input into the model, initiating a sampling process governed by $\rmd \bfx_t = \bfu_\theta(\bfx_t,\bfc)\rmd t$. 
Diffusion models and flow-based models are equivalent by $
    \alpha_t = \exp( \int_0^t f(u) \rmd u )$, and $
    \sigma_t = \sqrt{\int_0^t g_u^2 \exp( -2\int_0^u f(s)\rmd s ) \rmd u}$.
Practically, flow-based models are increasingly favored due to their conceptual simplicity, cultivating a series of cutting-edge text-to-image models such as Stable Diffusion v3.5~\cite{esser2024scaling} and Flux~\cite{flux}. Given that the primary focus of this paper is on state-of-the-art text-to-image models, we employ a flow-based formulation while still referring to them as diffusion models.


%% file: sec/method.tex
\section{Elucidating the Design Space of Sampling Acceleration}
\label{sec:elucidating}
\begin{figure*}[t]
    \centering
    \begin{subfigure}[b]{0.445\textwidth}
        \includegraphics[width=\textwidth]{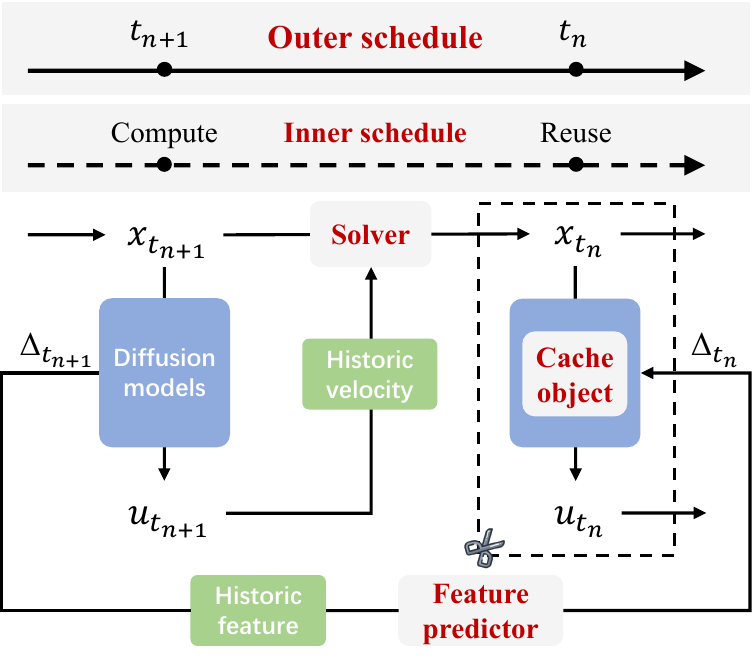}
        \caption{Unified framework of diffusion sampling.}
        \label{fig:components}
    \end{subfigure}
    \hfill
    \begin{subfigure}[b]{0.54\textwidth}
        \includegraphics[width=\textwidth]{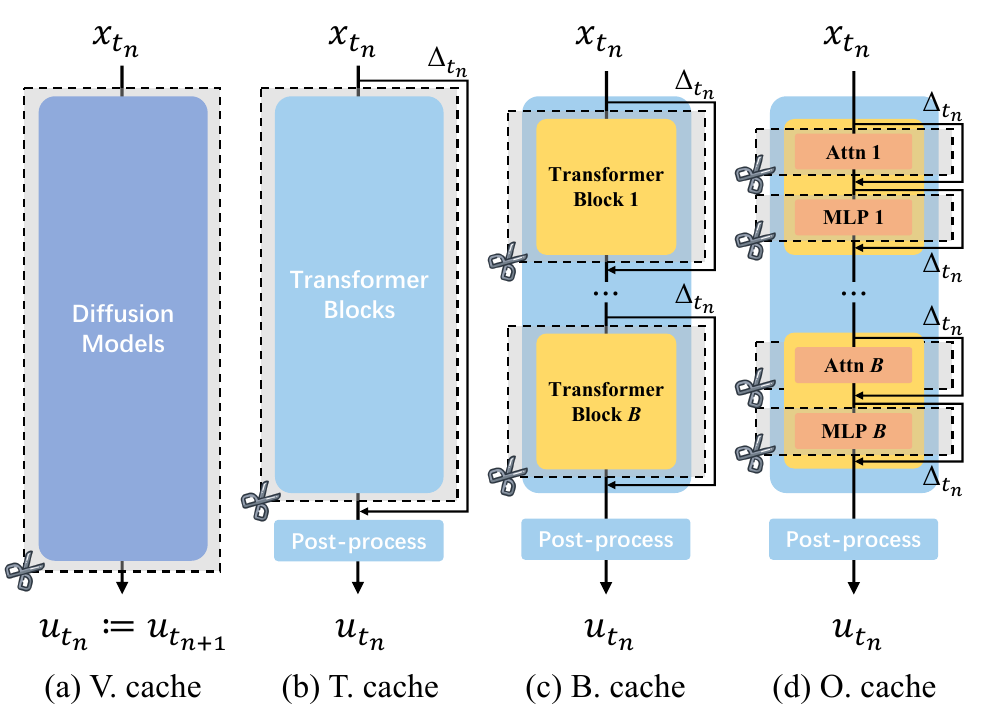}
        \caption{Four types of cache objects.}
        \label{fig:cache_object}
    \end{subfigure}
    \caption{(a) Diffusion sampling follows a predefined outer schedule $\{t_n\}_{n=0}^N$, while an inner schedule determines whether each step performs computation or feature reuse. A sample $\bfx_{t_{n+1}}$ transitions from $t_{n+1}$ to $t_n$ with the solver's update rule, utilizing stored historical velocities. Various types of cache objects can be selected, and a feature predictor is employed to estimate the current feature $\Delta_{t_n}$. This perspective also applies to U-Net architectures. (b) Four types of cache objects: velocity cache (V. cache), transformer cache (T. cache), block cache (B. cache), and operation cache (O. cache).}
    \label{fig:components_collect}
\end{figure*}

Diffusion models generate samples by simulating $\rmd \bfx_t = \bfu_\theta(\bfx_t,\bfc) \rmd t$ starting from Gaussian noise $\bfx_1$. 
A common discretization scheme is the Euler method, which reads $\bfx_{t_n} = \bfx_{t_{n+1}} + (t_n - t_{n+1}) \bfu_\theta(\bfx_{t_{n+1}},\bfc)$.
Due to the multi-step backward process, diffusion sampling is computationally expensive, and the growing parameter scales of modern text-to-image models further intensify the computational and memory demands~\cite{podell2024sdxl,esser2024scaling,flux}. While various training-free acceleration methods have been proposed to improve sampling efficiency, they are generally developed in isolation. In \Cref{fig:components}, we provide a unified perspective on these methods, including fast ODE solvers~\cite{karras2022edm,lu2022dpm,zhang2023deis,zhao2023unipc,zhou2023fast}, efficient time schedules~\cite{nichol2021improved,sabour2024align,chen2024trajectory,xue2024accelerating}, and feature caching~\cite{ma2024deepcache,wimbauer2024cache,li2023faster,liu2025reusing}. 
Below, we elucidate five key components within this unified design space.


\textbf{Solver}. Truncation errors in the Euler method accumulate rapidly when the number of sampling steps is small. 
To address this, multi-step higher-order solvers~\cite{lu2022dpm,zhang2023deis,zhao2023unipc} utilize historical velocities for more accurate sampling:
\begin{equation}
    \bfx_{t_n} = \bfx_{t_{n+1}} + (t_n - t_{n+1}) \sum_{i=1}^{O_S} \omega_{n+1,i}\bfu_\theta(\bfx_{t_{n+i}},\bfc),
\end{equation}
where $\{\omega_{n+1,i}\}_{i=1}^{O_S}$ are theoretically deduced coefficients to achieve an $O_S$-th order global truncation error.

\textbf{Outer schedule}. The sampling process of diffusion models starts from establishing the semantic structure and then refines the image details. An appropriate outer schedule ($0=t_0<t_1<\cdots<t_N=1$) is essential for high image quality, especially under limited sampling budgets.


Orthogonal to solvers and outer schedules, feature caching divides the sampling process into a series of ``compute-reuse'' cycles, where the stored features at previous compute steps are utilized in each reuse step to save computations. Feature caching contains three components:

\textbf{Inner schedule} determines whether to compute or reuse features at each $t_n$ of the given outer schedule. It is typically governed by a hyperparameter called \textit{cycle length}. A cycle length of $C$ indicates that each ``compute-reuse'' cycle consists of one compute step followed by $C-1$ reuse steps.

\textbf{Cache object} determines the content to be cached. Different cache objects impose different demands on both latency and VRAM. As shown in \Cref{fig:cache_object}, they can be categorized into four types:
(a) \textit{Velocity cache} bypasses all computations during reuse steps and outputs the velocity stored from the last compute step, effectively equivalent to removing all reuse steps from the outer schedule.
(b) \textit{Transformer cache} retains the residual outputs after processing through all the transformer blocks.
(c) \textit{Block cache} preserves the residual from each individual transformer block.
(d) \textit{Operation cache} delves into each block, storing the residuals of the attention and MLP modules, which account for the majority of computations within each transformer block.

\textbf{Feature predictor} operates similarly to solvers. It forecasts features required in the current ``reuse'' step by utilizing historical features from previous ``compute'' steps. An example of feature prediction with $C=2$ at the reuse step $t_n$ is given by
\begin{equation}
    \Delta_{t_n} = \sum_{i=1}^{O_F} \gamma_{n,i}\Delta_{t_{n+1+C(i-1)}},
\end{equation}
where ${O_F}$ is the maximum caching order and $\{\gamma_{n,i}\}_{i=1}^{O_F}$ are designed caching coefficients. A direct feature reuse is given by setting $O_F=1$ and $\gamma_{n,1}=1$, \ie, $\Delta_{t_n}=\Delta_{t_{n+1}}$.

\begin{figure*}[t]
    \centering
    \begin{subfigure}[b]{0.495\textwidth}
        \includegraphics[width=\textwidth]{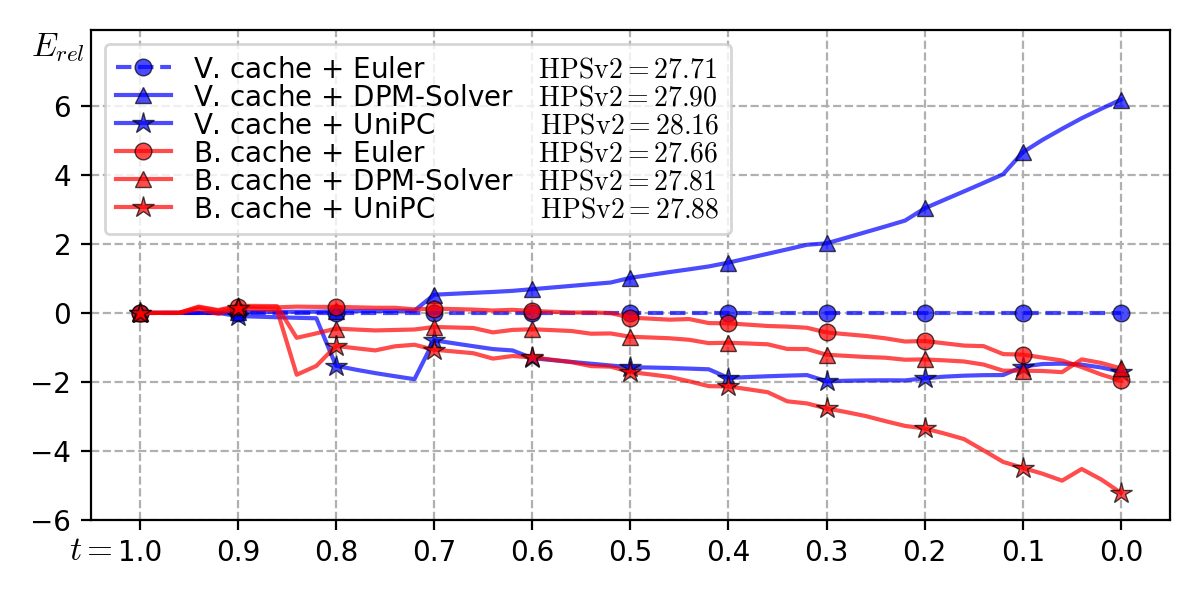}
        \caption{Solver.}
        \label{fig:compare_flux_solver}
    \end{subfigure}
    \hfill
    \begin{subfigure}[b]{0.495\textwidth}
        \includegraphics[width=\textwidth]{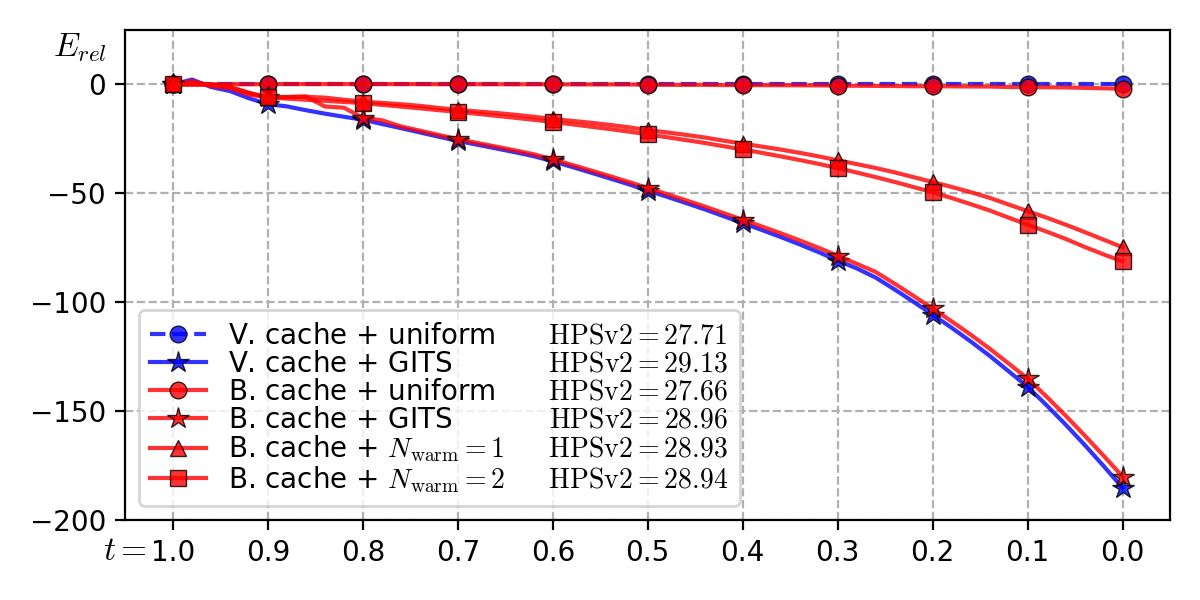}
        \caption{Outer schedule.}
        \label{fig:compare_flux_outer}
    \end{subfigure}
    \hfill
    \begin{subfigure}[b]{0.495\textwidth}
        \includegraphics[width=\textwidth]{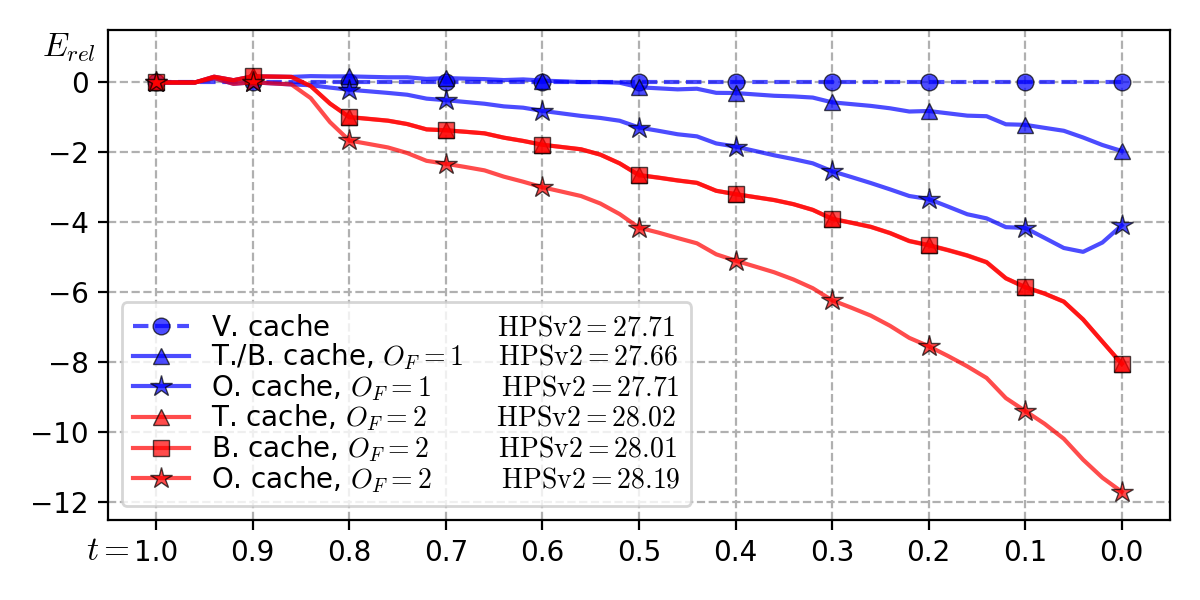}
        \caption{Cache object.}
        \label{fig:compare_flux_object}
    \end{subfigure}
    \hfill
    \begin{subfigure}[b]{0.495\textwidth}
        \includegraphics[width=\textwidth]{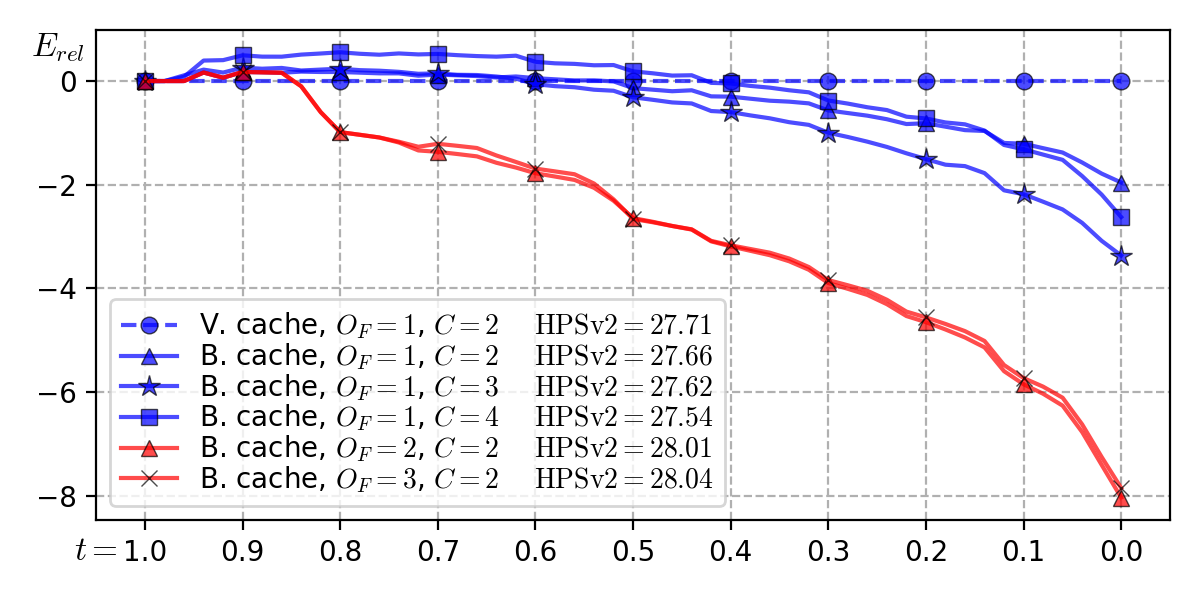}
        \caption{Inner schedule and feature predictor.}
        \label{fig:compare_flux_pred_inner}
    \end{subfigure}
    \caption{Comprehensive experiments examining the impact of each acceleration method on sampling acceleration. The experiments are conducted on Flux.1-Dev~\cite{flux}. The outer schedule is identified as the most influential factor (see the vertical scale range).}
    \label{fig:compare_flux}
\end{figure*}

\section{Analyzing Training-Free Sampling Acceleration}
\label{sec:analyzing}

In this section, we conduct comprehensive experiments to analyze the contribution of each component in training-free acceleration methods, examining their effects on both the \textit{sampling trajectory} $\{\bfx_{t_n}\}_{n=1}^N$ and overall performance.

\textbf{Settings}. As our primary focus is sampling acceleration, we evaluate models under a 10-step compute setting, where state-of-the-art text-to-image models struggle to maintain high visual quality. To integrate feature caching methods into a unified perspective, we employ $N=20$ sampling steps with a cycle length of $C=2$, resulting in 10 total compute steps. 
Unless otherwise specified, we employ Euler discretization and a uniform outer schedule, both of which are default settings in recent text-to-image models~\cite{rombach2022ldm,podell2024sdxl,esser2024scaling,flux}. 
For cache configuration, we implement a block cache with a maximum caching order of $O_F=1$ by default. When $O_F>1$, we employ the feature predictor from TaylorSeers~\cite{liu2025reusing}, which calculates adaptive caching coefficients via Taylor expansion.

\textbf{Metrics}. Given a baseline sampling trajectory $\{\hat{\bfx}_{t_n}\}_{n=1}^{N_b}$ with $N_b$ sampling steps and a reference sampling trajectory $\{\bar{\bfx}_{t_n}\}_{n=1}^{N_r}$ with $N_r$ sampling steps, the \textit{relative trajectory error} of a target sampling trajectory $\{\bfx_{t_n}\}_{n=1}^{N}$ is defined as
\begin{equation}
    E_{rel}(t) = \lVert \bar{\bfx}_t - \bfx_t \rVert_2 - \lVert \hat{\bfx}_t - \bfx_t \rVert_2.
\end{equation}
The relative formulation facilitates clearer visualization. The baseline and reference trajectories are generated using velocity cache with $N_b=10$ and $N_r=50$ steps ($C=1$), respectively, corresponding to standard sampling without acceleration. 
A negative $E_{rel}$ indicates that the target trajectory is closer to the reference trajectory than the baseline trajectory.
As the $L_2$ distance does not adequately correlate with visual quality, we also report HPSv2 (Human Preference Score v2~\cite{wu2023human}), which better predicts human perceptual preferences, to evaluate the visual quality of decoded samples on the target trajectories. 
All the metrics are averaged over 200 prompts. Results on Flux.1-Dev~\cite{flux} are shown in \Cref{fig:compare_flux}. Results for Stable Diffusion v3.5 are included in Appendix \ref{subsec:app_quantitiative}.

\textbf{Solver}. In addition to the Euler discretization, we further incorporate two fast solvers, namely, DPM-Solver~\cite{lu2022dpm} and UniPC~\cite{zhao2023unipc}, both set with a maximum solver order of $O_S=2$. 
The results in \Cref{fig:compare_flux_solver} demonstrate that although these solvers exhibit distinct behaviors in terms of relative trajectory error, they all enhance image quality.

\textbf{Outer schedule}. 
Beyond the default uniform outer schedule, we re-implement GITS~\cite{chen2024trajectory}, an effective schedule obtained utilizing dynamic programming by approximating global truncation error with accumulated local errors.
As illustrated in~\Cref{fig:compare_flux_outer}, GITS significantly improves both relative trajectory error and image quality.
Inspecting the resulting GITS schedule, we observe a dense allocation of compute steps in the early sampling stage.
To better understand this phenomenon, we adopt a simple reallocating strategy for block cache, where the ``compute-reuse'' cycles initiate after $N_{\mathrm{warm}}$ consecutive compute steps while omitting the last few compute steps to maintain a total of 10 compute steps for fair comparison. 
This warm-up strategy reallocates the computational effort toward earlier sampling stages, yielding considerable performance gains. In contrast, shifting compute steps toward the final stage degrades performance. Similar results are observed on Stable Diffusion 3.5, as detailed in Appendix \ref{subsec:app_quantitiative}. 
Therefore, we conclude that the early sampling stage is crucial for state-of-the-art text-to-image models and merits prioritized resource allocation.

\textbf{Cache object}. We compare four representative cache objects that progressively extend into deeper network modules. With a maximum caching order of $O_F=1$, the transformer cache and block cache are equivalent. Results in \Cref{fig:compare_flux_object} demonstrate that naive caching ($O_F=1$) offers limited benefit, while higher caching orders yield modest improvements, particularly for deeper levels. 
However, deeper caching also increases latency and VRAM demands. 
For instance, to generate a single image using Flux.1-Dev, velocity cache requires 9.85 seconds and 36.32 GB of VRAM. In contrast, the deepest operation cache incurs a 6$\%$ to 14$\%$ latency increase and requires 38.48 GB to 42.78 GB of VRAM, even surpassing the memory needed for a doubled batch size with velocity cache (38.88 GB).

\textbf{Inner schedule and feature predictor}. We conduct ablation studies on the inner schedule and feature predictor by adjusting the cycle length $C$ and maximum caching order $O_F$, respectively. 
As shown in \Cref{fig:compare_flux_pred_inner}, adjusting the inner schedule does not improve performance, whereas a higher caching order slightly enhances performance, albeit at the cost of increased latency and VRAM requirements.

All experiments above are replicated on Stable Diffusion v3.5~\cite{esser2024scaling}, with results detailed in Appendix \ref{subsec:app_quantitiative}. 
The trends for solver and outer schedule remain consistent. However, while feature caching methods may enhance image quality on Flux~\cite{flux} models, no performance gains are observed on Stable Diffusion v3.5~\cite{esser2024scaling}. 

\textbf{Summary of analyses}. To summarize, our key findings on training-free sampling acceleration are as follows:
\begin{itemize}
    \item {\em Outer schedule} impacts the performance most: allocating more compute steps to the early sampling stages substantially enhances performance.
    \item Higher-order {\em solvers} consistently yield slight improvements in image quality.
    \item The effectiveness of the {\em feature predictor} varies across text-to-image models.
    \item Deeper {\em cache object} without higher-order feature predictor offers no additional benefit.
    \item Increasing the cycle length for the {\em inner schedule} offers no performance gain.
\end{itemize}
Given the significant performance gains achieved by optimizing the outer schedule, we focus on further enhancements to this aspect in the subsequent sections.

\section{Improving Training-Free Sampling Acceleration}
\label{sec:methods}
\subsection{Uniform Outer Schedule is Suboptimal}
\label{subsec:flawed}

\begin{figure*}[t]
    \centering
    \includegraphics[width=\textwidth]{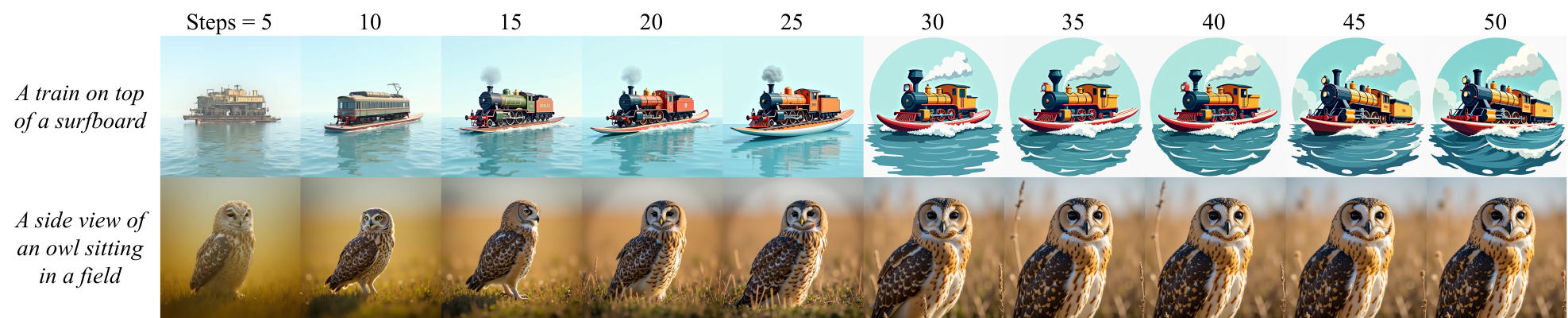}
    \caption{Text-to-image generation with Flux.1-Dev~\cite{flux} using the default uniform outer schedule. As the number of sampling steps increases, the image structure continues to change and only stabilizes after around 30 steps.}
    \label{fig:flawed}
\end{figure*}

A widely recognized principle in the sampling dynamics of diffusion models is that early steps define semantic content, while later steps refine visual details~\cite{chen2023geometric}. However, we observe slow structural convergence for text-to-image models with the default uniform outer schedule. As shown in \Cref{fig:flawed}, image structure fluctuates notably before stabilizing at around 30 steps.

To understand this phenomenon, we first extend the findings on the geometric regularity in diffusion models, previously explored for VE SDEs~\cite{chen2023geometric,chen2024trajectory}, to the flow-based framework utilized by state-of-the-art text-to-image models. We conduct Principal Component Analysis (PCA) on 100 sampling trajectories, each with 100 sampling steps. The results presented in \Cref{fig:regularity_flux} reveal strong trajectory regularity, with an average explained variance exceeding 99$\%$ using the top \textit{three} principal components. Therefore, we can closely characterize each trajectory in the subspace spanned by these principal components, which is referred to as the \textit{projected sampling trajectory}. We further observe that these projected sampling trajectories begin with a high-curvature phase (Figures \ref{fig:regularity_flux}, \ref{fig:curv_tors_flux}), which requires smaller step sizes to reduce truncation errors. This is consistent with the empirically successful approaches such as GITS~\cite{chen2024trajectory} and the warm-up strategy. In contrast, a coarse uniform outer schedule that underrepresents this early phase tends to induce prolonged structural instability in the generated images. Motivated by these geometric insights and experimental evidence, we introduce an improved scheduling strategy tailored for modern text-to-image diffusion models.

\subsection{Constant Total Rotation Schedule}
\label{subsec:ours}

The core mechanism of the GITS algorithm lies in approximating the global truncation error by accumulating local errors. In contrast, our approach advances this idea by explicitly incorporating the underlying geometric properties of sampling trajectories, in particular, curvature and torsion in the three-dimensional subspace, to construct a more computationally efficient outer scheduling strategy.

\textbf{Curvature and torsion}. A spatial curve that is at least twice differentiable in three-dimensional Euclidean space $\mathbb{R}^3$ can be characterized by the {\em Frenet–Serret formulas}~\cite{pressley2010elementary,chen2025geometric}, describing the evolution of an orthonormal frame formed by the tangent $\mathbf{T}$, normal $\mathbf{N}$, and binormal $\mathbf{B}$ vectors along the curve. These formulas quantify the geometric properties of the curve through two scalar functions: the curvature $\kappa$, measuring the change rate of the tangent direction, and the torsion $\tau$, capturing the rate at which the curve deviates from its osculating plane.
Consider $\bfr(t)$ as a projected sampling trajectory in $\bbR^3$. Since different trajectories may trace the same geometrical shape at different speeds, we adopt an arc-length parameterization to ensure a speed-independent representation. Specifically, assuming that $\dot{\bfr} \neq 0$, we define $\bfr(s)=\bfr(t(s))$ where $s(t)=\int_0^t\lVert \dot{\bfr}(u) \rVert \rmd u$ is the arc length. With a non-degenerate curve $\bfr(s)$, the Frenet-Serret formulas are:
\begin{equation}
    \frac{\rmd \bfT}{\rmd s} = \kappa \bfN, \quad
    \frac{\rmd \bfN}{\rmd s} = -\kappa \bfT + \tau \bfB, \quad
    \frac{\rmd \bfB}{\rmd s} = -\tau \bfN. 
\end{equation}

Curvature $\kappa=\lVert \frac{\rmd \bfT}{\rmd s} \rVert$ measures how sharply a curve bends at a given point, while torsion $\tau=-\bfN\cdot \frac{\rmd \bfB}{\rmd s}$ measures the rate at which the curve deviates from being planar, capturing its twist into the third dimension~\cite{pressley2010elementary}. The fundamental theorem of space curves states that curvature and torsion uniquely determine the shape of a regular curve in three-dimensional space (up to rigid motion), a principle that underpins many practical applications in geometry, physics, and computer vision.

\begin{figure*}[t]
    \centering
    \begin{subfigure}[b]{0.33\textwidth}
        \includegraphics[width=\textwidth]{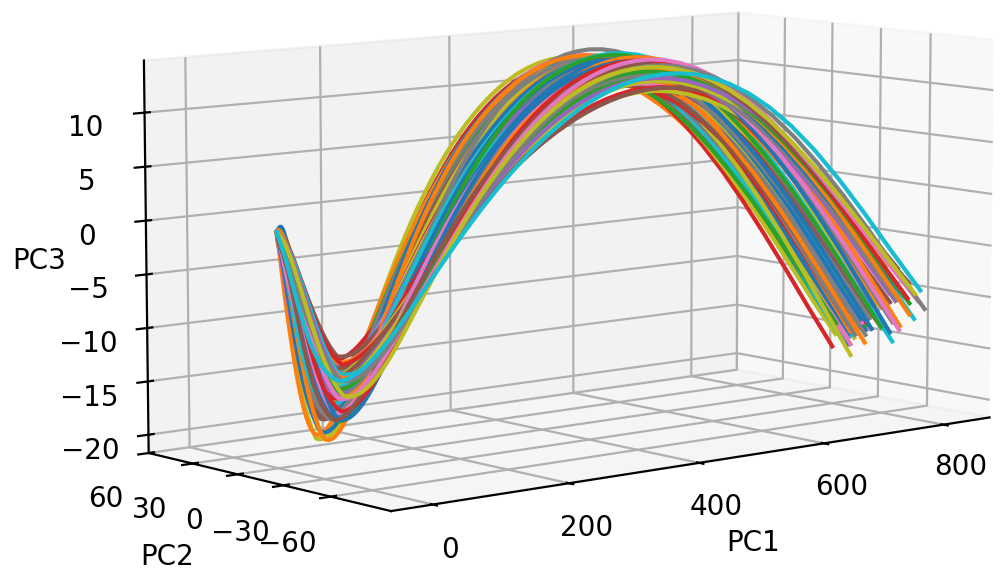}
        \caption{Trajectory regularity observed on Flux.1-Dev.}
        \label{fig:regularity_flux}
    \end{subfigure}
    \hfill
    \begin{subfigure}[b]{0.29\textwidth}
        \includegraphics[width=\textwidth]{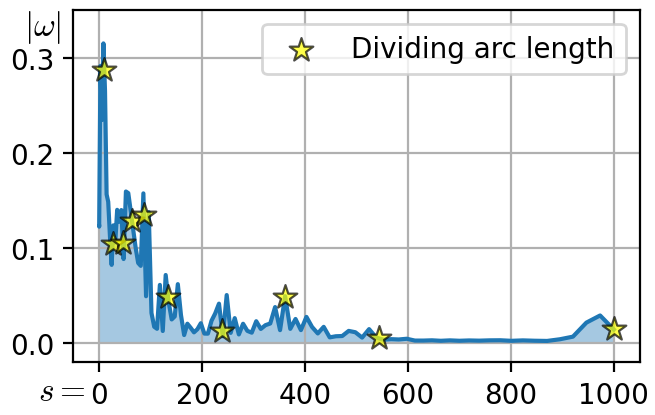}
        \caption{Selected dividing arc lengths for \ourName.}
        \label{fig:selected}
    \end{subfigure}
    \hfill
    \begin{subfigure}[b]{0.35\textwidth}
        \includegraphics[width=\textwidth]{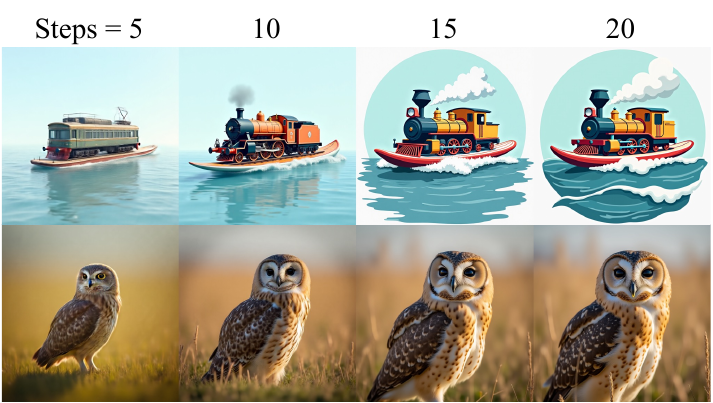}
        \caption{\ourName yields faster structural convergence.}
        \label{fig:ours_qualitative}
    \end{subfigure}
    \caption{We extend the insights of sampling regularity from diffusion models to flow-based models and propose \ourName by utilizing the geometric properties of sampling trajectories. (a) Visualization of 100 sampling trajectories generated by Flux.1-Dev reveals a strong trajectory regularity. All sampling trajectories starting from Gaussian noise are uniformly shifted to the origin. (b) A 10-step example of our proposed \ourName, which ensures a constant total rotation change $\int_0^S|\omega(s)|\rmd s$ along the sampling trajectory. (c) \ourName yields considerably faster structural convergence compared with the uniform outer schedule (\Cref{fig:flawed}).}
    \label{fig:regularity}
\end{figure*}

\textbf{Total rotation}. The uniform outer schedule overlooks these geometric properties, adopting excessively large step sizes during the initial stage of sampling, where both curvature and torsion are high. Consequently, this leads to a prolonged structural instability in the images before convergence. To address this, we utilize these geometric properties to derive an improved, efficient outer schedule. Within the Frenet-Serret framework, the angular velocity vector of a space curve, known as the \textit{Darboux vector}, is defined by $\boldsymbol{\omega}=\tau\bfT+\kappa\bfB$~\cite{farouki2008pythagorean}. Its magnitude, $\lvert \boldsymbol{\omega} \rvert=\sqrt{\kappa^2 + \tau^2}$, represents the instantaneous rotation rate, which is usually called the \textit{total curvature} of the curve. Therefore, we define the \textit{total rotation} of a segment as
\begin{equation}
    \Theta_{[s_1, s_2]} = \int_{s_1}^{s_2}\sqrt{\kappa^2(s) + \tau^2(s)}\rmd s.
\end{equation}

\textbf{TORS schedule}. Leveraging the geometric insights of the projected sampling trajectory, we allocate more compute steps to regions where the curvature and torsion are high and fewer compute steps otherwise. To achieve this, we propose adjusting step sizes to maintain a constant change in total rotation. 
Specifically, we compute the average curvature $\kappa(s)$ and torsion $\tau(s)$ over a fine-grained arc-length grid across 100 projected sampling trajectories. 
These statistics, defined over $s\in[0,S]$, serve as indicators of the model's geometric behavior. 
Utilizing these statistics, we partition the trajectory into $N$ segments, each representing an equal share of the total rotation, \ie, $\frac{\Theta_{[0, S]}}{N}$. 
Based on the selected dividing arc lengths (see \Cref{fig:selected}), our proposed efficient time schedule is obtained by mapping each arc-length $s$ back to its corresponding timestamp $t$. We name this scheduling approach the \textit{cons\textbf{t}ant t\textbf{o}tal \textbf{r}otation \textbf{s}chedule} (\ourName), reflecting its strategic focus on maintaining a constant change in total rotation throughout sampling. 
The full algorithm is presented in \Cref{alg:tors_generation}.
As shown in \Cref{fig:ours_qualitative}, \ourName yields considerably faster convergence on image quality compared with the uniform outer schedule.


\textbf{Justification for total rotation}. 
Define the Frenet-Serret basis $\mathbf{F}(s)=[\bfT(s),\bfN(s),\bfB(s)]$.
By the Frenet-Serret formulas, $\lVert \mathbf{F}'(s) \rVert_F = \sqrt{2\kappa^2+2\tau^2} = \sqrt{2}\,\lvert \boldsymbol{\omega}(s) \rvert$, 
where $\lvert \boldsymbol{\omega} \rvert$ characterizes the instantaneous rotation rate of the orthonormal frame.
To assess its role in discretization accuracy, we examine the local truncation error of Euler steps along $\bfr(s)$. A Taylor expansion in the Frenet-Serret frame yields
\begin{equation}
\label{eq:error_decomp}
\boldsymbol{\epsilon} = \frac{h^{2}}{2}\kappa\,\bfN + \frac{h^{3}}{6}\kappa\tau\,\bfB + O(h^{4}),
\end{equation}
neglecting subdominant terms as $\kappa$ is dominated by $\tau$ along diffusion trajectories (\Cref{fig:curv_tors_flux}).
This decomposition reveals two geometric error sources: normal bending ($\kappa$, $O(h^{2})$) and binormal twisting ($\kappa\tau$, $O(h^{3})$).
While classical convergence analysis ($h \to 0$) ignores higher-order terms, in the few-step regime with large $h$, the $O(h^{3})$ binormal error becomes non-negligible---potentially comparable to the $O(h^{2})$ normal error as their ratio scales with $h\tau/3$.
The total rotation rate $\lvert \boldsymbol{\omega} \rvert=\sqrt{\kappa^{2}+\tau^{2}}$ naturally couples these two dominant error sources into a single scalar, serving as a principled indicator for adaptive discretization.
Equidistributing total rotation $\Theta$ thus ensures that the regions of high bending or twisting receive proportionally denser sampling steps, providing unified control over the leading geometric errors.

\renewcommand{\arraystretch}{0.9}
\begin{table*}[t]
  \caption{Results on text-to-image generation. Our proposed \ourName achieves state-of-the-art results among sampling acceleration methods.}
  \label{tab:exp}
  \centering
  \fontsize{8}{10}\selectfont
  \begin{subtable}[b]{0.47\textwidth}
    \caption{Results on Flux.1-Dev.}
    \label{subtab:exp_flux}
    \centering
    \begin{tabular}{lcccc}
        \toprule
        Methods & IR & CS & AS & HPSv2 \\
        \midrule
        \multicolumn{5}{l}{\fontsize{6.5}{1}\selectfont\textbf{Baselines}} \\
        \midrule
        50 steps            & 0.96 & 30.61 & 5.73 & 30.15 \\
        20 steps            & 0.93 & 30.76 & 5.68 & 29.30 \\
        10 steps            & 0.71 & 30.08 & 5.53 & 27.70 \\
        \midrule
        \multicolumn{5}{l}{\fontsize{6.5}{1}\selectfont\textbf{Sampling acceleration methods} (10 steps)} \\
        \midrule
        \rowcolor{orange!15}FORA~\cite{selvaraju2024fora}       & 0.71 & 30.32 & 5.53 & 27.71 \\
        \rowcolor{orange!15}TaylorSeers~\cite{liu2025reusing}   & 0.77 & 30.38 & 5.58 & 28.19 \\
        \rowcolor{teal!20}DPM-Solver~\cite{lu2022dpm}         & 0.73 & 30.27 & 5.58 & 27.98 \\
        \rowcolor{teal!20}UniPC~\cite{zhao2023unipc}          & 0.78 & 30.41 & 5.58 & 28.16 \\
        \rowcolor{blue!10}TPDM~\cite{ye2025schedule}      & 0.76 & 30.07 & 5.69 & 28.82 \\
        \rowcolor{blue!10}GITS~\cite{chen2024trajectory}      & 0.90 & 30.53 & 5.65 & 29.13 \\
        \rowcolor{blue!10}\textbf{\ourName (ours)}            & \textbf{0.97} & \textbf{30.97} & \textbf{5.71} & \textbf{29.30} \\
        \bottomrule
    \end{tabular}
  \end{subtable}
  \quad
  \begin{subtable}[b]{0.47\textwidth}
    \caption{Results on Stable Diffusion 3.5 medium.}
    \label{subtab:exp_sd3}
    \centering
    \begin{tabular}{lcccc}
        \toprule
        Methods & IR & CS & AS & HPSv2 \\
        \midrule
        \multicolumn{5}{l}{\fontsize{6.5}{1}\selectfont\textbf{Baselines}} \\
        \midrule
        50 steps            & 0.97 & 33.26 & 5.37 & 28.64 \\
        20 steps            & 0.94 & 33.32 & 5.36 & 27.95 \\
        10 steps            & 0.55 & 32.77 & 5.23 & 25.31 \\
        \midrule
        \multicolumn{5}{l}{\fontsize{6.5}{1}\selectfont\textbf{Sampling acceleration methods} (10 steps)} \\
        \midrule
        \rowcolor{orange!15}FORA~\cite{selvaraju2024fora}       & 0.56 & 32.81 & 5.27 & 25.40 \\
        \rowcolor{orange!15}TaylorSeers~\cite{liu2025reusing}   & 0.54 & 32.73 & 5.22 & 25.25 \\
        \rowcolor{teal!20}DPM-Solver~\cite{lu2022dpm}         & 0.61 & 32.88 & 5.26 & 25.73 \\
        \rowcolor{teal!20}UniPC~\cite{zhao2023unipc}          & 0.69 & 32.88 & 5.29 & 26.43 \\
        \rowcolor{blue!10}TPDM~\cite{ye2025schedule}      & 0.60 & 33.10 & 5.26 & 25.47 \\
        \rowcolor{blue!10}GITS~\cite{chen2024trajectory}      & 0.75 & 32.72 & 5.25 & 25.89 \\
        \rowcolor{blue!10}\textbf{\ourName (ours)}            & \textbf{0.86} & \textbf{33.13} & \textbf{5.33} & \textbf{26.90} \\
        \bottomrule
    \end{tabular}
  \end{subtable}
\end{table*}

%% file: sec/exp.tex
\section{Experiments}
\label{sec:exp}

\renewcommand{\arraystretch}{0.8}
\begin{table}[h]
    \centering
    \caption{GenEval evaluation on Flux.1-dev.}
    \label{tab:geneval}
    \fontsize{8}{10}\selectfont
    \begin{tabular}{lc|cccccc|c}
        \toprule
        \textbf{Model} & \textbf{Steps} & \makecell{\textbf{Single}\\\textbf{Object}} & \makecell{\textbf{Two}\\\textbf{Object}} & \textbf{Counting} & \textbf{Colors} & \makecell{\textbf{Attribute}\\\textbf{Binding}} & \textbf{Position} & \textbf{Overall}$\uparrow$ \\
        \midrule
        Baseline & 50    & 0.99 & 0.81 & \textbf{0.69} & \textbf{0.81} & 0.41 & 0.20 & 0.65 \\
        Baseline & 10    & 0.96 & 0.71 & 0.59 & 0.69 & 0.28 & 0.20 & 0.57 \\
        \textbf{TORS (ours)} & 10    & \textbf{1.00} & \textbf{0.82} & 0.67 & 0.80 & \textbf{0.49} & \textbf{0.22} & \textbf{0.67} \\
        \bottomrule
    \end{tabular}
\end{table}

\begin{figure}[t]
  \centering
  \begin{minipage}[b]{0.49\textwidth}
    \centering
    \includegraphics[width=\linewidth]{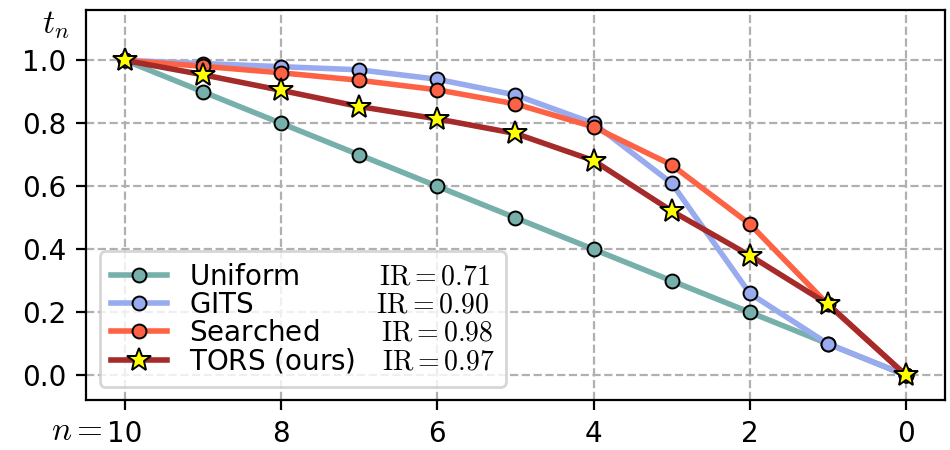}
    \caption{Visualizing different outer schedules. \ourName achieves comparable performance with the searched optimal schedule.}
    \label{fig:search_flux}

    \includegraphics[width=\linewidth]{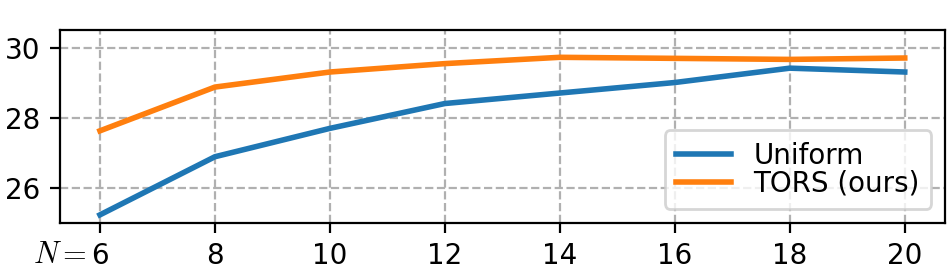}
    \caption{Ablation on NFE evaluated by HPSv2. \ourName largely outperforms the default schedule in few-step generation.}
    \label{fig:ablation_nfe}
  \end{minipage}
  \hfill
  \begin{minipage}[b]{0.49\textwidth}
    \centering
    \includegraphics[width=\linewidth]{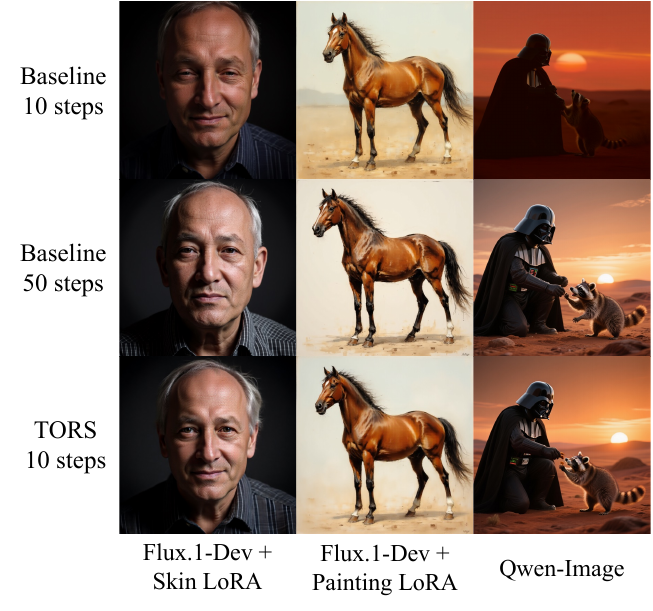}
    \caption{\ourName schedules generated for Flux generalizes well on unseen LoRA-fine-tuned variants as well as the 20B Qwen-Image model.}
    \label{fig:transferability}
  \end{minipage}
\end{figure}

In this section, we conduct comprehensive experiments to demonstrate the superiority of our proposed \ourName and examine its compatibility with the aforementioned sampling acceleration methods. Sample quality is measured by Image Reward (IR)~\cite{xu2023imagereward}, CLIP Score (CS)~\cite{radford2021learning}, Aesthetic Score (AS)~\cite{schuhmann2022laion} and HPSv2~\cite{wu2023human} on the DrawBench dataset~\cite{saharia2022photorealistic}. Two recent text-to-image diffusion/flow-based models, namely, Flux.1-Dev~\cite{flux} and Stable Diffusion 3.5 medium~\cite{esser2024scaling} are employed with the guidance scale of 7, unless otherwise specified. All experiments are conducted using up to 4 NVIDIA RTX A6000 GPUs.

\subsection{Performance Comparison}
\label{subsec:exp_ours}
We compare our proposed \ourName with cutting-edge training-free sampling acceleration methods. For feature caching, we report results using FORA~\cite{selvaraju2024fora} and TaylorSeers~\cite{liu2025reusing}, which are representative methods tailored for transformer-based diffusion models. In terms of solvers, we include two advanced fast solvers, namely DPM-Solver~\cite{lu2022dpm} and UniPC~\cite{zhao2023unipc}. For the outer schedule, we compare our scheduling strategy with GITS~\cite{chen2024trajectory}, an efficient training-free time schedule developed through dynamic programming. 
We further compare \ourName with representative training-based scheduling methods~\cite{tong2024learning,ye2025schedule}. While approaches like LD3~\cite{tong2024learning} demonstrate promising results in low-resolution regimes, they are computationally prohibitive due to the requirement of backpropagating gradients through the entire sampling chain. Therefore, we re-implement TPDM~\cite{ye2025schedule} that sidesteps the unrolled computational graph via reinforcement learning.
Implementation details are included in Appendix \ref{subsec:app_implementation}.
The results of the performance comparison are shown in \Cref{tab:exp}, where our proposed \ourName consistently outperforms existing sampling acceleration methods under a 10-step compute budget. Notably, \ourName achieves nearly $5\times$ acceleration on the Flux model, closely approaching the performance of the 50-step baseline. Our method delivers visually superior results, as shown in \Cref{fig:teaser} 
and Appendix C.5.
Furthermore, evaluated on GenEval~\cite{ghosh2023geneval} (\Cref{tab:geneval}), \ourName with 10 steps significantly outperforms the 10-step baseline and even surpasses the 50-step baseline in overall score.

To further illustrate the superiority of \ourName, we design a complex outer schedule parameterized by $\alpha,\beta$ and $p$, aiming to encompass a variety of shapes:
\begin{equation}
    t_n=\left(1 - I_{\frac{N-n}{N}}(\alpha,\beta)\right)\left(\frac{n}{N}\right)^p,
\end{equation}
where $I_t(\alpha,\beta)=\frac{B_t(\alpha,\beta)}{B_1(\alpha,\beta)}$ and $B_t(\alpha,\beta)=\int_0^tx^{\alpha-1}(1-x)^{\beta-1}\rmd x$ is called the \textit{incomplete beta function}. We employ Bayesian optimization to search for an optimal outer schedule within the parameter space, which minimizes the distance between pairs of 10-step and 50-step samples. 
We conduct 1000 iterations of parameter searching, obtaining an optimal set of parameters as $\alpha=6.23, \beta=1.34$ and $p=0.18$.
In \Cref{fig:search_flux}, we visualize various 10-step outer schedules. Our proposed \ourName achieves performance comparable to the searched optimal outer schedule in terms of Image Reward, demonstrating the effectiveness of our scheduling strategy. Moreover, as shown in \Cref{fig:ablation_nfe}, \ourName schedules largely outperform the default uniform schedule in few-step generation.


\begin{table}[t]
  \centering
  \begin{minipage}[t]{0.55\textwidth}
    \centering
    \caption{Quantitative results for adaptability.}
    \label{tab:ablation_archi}
    \renewcommand{\arraystretch}{0.92}
    \fontsize{8}{10}\selectfont
    \begin{tabular}{lccccc}
        \toprule
        Model & $N$ & IR & CS & AS & HPSv2 \\ 
        \midrule
        \rowcolor[gray]{0.9}Flux + Skin LoRA        & 50 & 0.70 & 29.86 & 5.50 & 28.91 \\
        Flux + Skin LoRA        & 10 & 0.42 & 29.35 & 5.41 & 26.59 \\
        + \textbf{TORS (ours)}  & 10 & \textbf{0.67} & \textbf{30.15} & \textbf{5.57} & \textbf{28.24} \\
        \midrule
        \rowcolor[gray]{0.9}Flux + Painting LoRA    & 50 & 0.99 & 30.42 & 5.79 & 29.26 \\
        Flux + Painting LoRA        & 10 & 0.81 & 30.15 & 5.79 & 27.56 \\
        + \textbf{TORS (ours)}      & 10 & \textbf{0.98} & \textbf{31.03} & \textbf{5.81} & \textbf{28.52} \\
        \midrule
        \rowcolor[gray]{0.9} Qwen-Image  & 50 & 1.18 & 33.78 & 5.57 & 30.80 \\
        Qwen-Image                      & 10 & 0.91 & 33.18 & 5.46 & 27.10 \\
        + \textbf{TORS (ours)}          & 10 & \textbf{1.12} & \textbf{33.76} & \textbf{5.55} & \textbf{29.28} \\
        \bottomrule
    \end{tabular}
  \end{minipage}
  \hfill
  \begin{minipage}[t]{0.4\textwidth}
    \centering
    \caption{Ablation study on guidance scales. \ourName schedules generated with different guidance scales remain effective across different inference guidance scales.}
    \label{tab:ablation_cfg}
    \renewcommand{\arraystretch}{0.9}
    \fontsize{8}{10}\selectfont
    \begin{tabular}{cccccc}
        \toprule
        \multirow{2}{*}{\parbox{1.5cm}{\centering Inference \\ CFG}} & \multirow{2}{*}{Base.} & \multicolumn{4}{c}{TORS CFG} \\ 
        \cmidrule(lr){3-6}
        & & 3 & 5 & 7 & 9 \\ \midrule
        3   & 0.88 & 0.93 & 0.93 & 0.95 & 0.97 \\
        5   & 0.72 & 0.92 & 0.92 & 0.95 & 0.92 \\
        7   & 0.71 & 0.96 & 0.93 & 0.97 & 0.94 \\
        9   & 0.78 & 0.91 & 0.87 & 0.91 & 0.92 \\
        \bottomrule
    \end{tabular}
  \end{minipage}
\end{table}

\renewcommand{\arraystretch}{0.9}
\begin{table*}[t]
  \caption{Ablation studies evaluating the robustness of \ourName schedules on prompt distribution, sample sizes and step counts.}
  \label{tab:ablations}
  \centering
  \fontsize{8}{10}\selectfont
  \begin{subtable}[b]{0.42\textwidth}
    \caption{Prompt distribution.}
    \label{subtab:ablation_prompt}
    \centering
    \begin{tabular}{lcccc}
      \toprule
      Prompt set & IR & CS & AS & HPSv2 \\ 
      \midrule
      Anime       & 0.94 & 31.00 & 5.71 & 29.57 \\
      Concept     & 0.94 & 31.04 & 5.69 & 29.35 \\
      Paintings   & 0.95 & 30.96 & 5.70 & 29.37 \\
      Photo       & 0.96 & 31.01 & 5.69 & 29.31 \\
      \bottomrule
    \end{tabular}
  \end{subtable}
  \hfill
  \begin{subtable}[b]{0.27\textwidth}
    \caption{Sample sizes.}
    \label{subtab:ablation_samples}
    \centering
    \centering
    \begin{tabular}{lccc}
      \toprule
      $\#$Samples & IR & HPSv2 \\ \midrule
      25  & 0.96 & 29.41 \\
      50  & 0.96 & 29.43 \\
      75  & 0.95 & 29.37 \\
      100 & 0.97 & 29.30 \\ \bottomrule
    \end{tabular}
  \end{subtable}
  \hfill
  \begin{subtable}[b]{0.27\textwidth}
    \caption{Steps counts.}
    \label{subtab:ablation_steps}
    \centering
    \begin{tabular}{lccc}
      \toprule
      $\#$Steps & IR & HPSv2 \\ \midrule
      25  & 0.81 & 28.23 \\
      50  & 0.91 & 28.61 \\
      75  & 0.94 & 29.03 \\
      100 & 0.97 & 29.30 \\ \bottomrule
    \end{tabular}
  \end{subtable}
\end{table*}


\subsection{Ablation studies}
\label{subsec:ablation}
To compute the geometric statistics required for \ourName scheduling, we generate 100 sampling trajectories, each with 100 sampling steps, using text prompts randomly sampled from the MS-COCO dataset~\cite{lin2014microsoft}. The computational overhead is approximately three and one A6000 hours for Flux.1-Dev and Stable Diffusion 3.5 medium, respectively. Notably, these pre-computed statistics enable the instantaneous generation of \ourName schedules for an arbitrary number of sampling steps with negligible overhead. 
To further demonstrate the efficiency and effectiveness, we provide additional experiments that underscore the adaptability, robustness, and transferability of our \ourName schedules.

\textbf{Adaptability}. We verify the adaptability of our method in \Cref{fig:transferability} and \Cref{tab:ablation_archi} by \textbf{directly applying} the \ourName schedules, pre-computed for Flux, to different LoRA-fine-tuned variants and distinct architectures such as Qwen-Image~\cite{wu2025qwenimagetechnicalreport}. Notably, our method exhibits strong generalization ability to unseen models, maintaining high image quality under constrained sampling budgets.

\textbf{Robustness}. In \Cref{tab:ablation_cfg}, we conduct an ablation study on guidance scales by generating \ourName schedules under various configurations (referred to as TORS CFG) and evaluating their performance with different inference CFGs. The results demonstrate that our method maintains its efficacy regardless of the guidance scale employed. Furthermore, \Cref{tab:ablations} provides comprehensive ablations on the prompt distributions, sample sizes, and step counts used for schedule generation. Our method is largely insensitive to the specific prompt set and the number of samples. However, maintaining a sufficient number of sampling steps during profiling is essential for capturing accurate geometric statistics of the trajectory.

\begin{figure*}[t]
    \centering
    \includegraphics[width=\textwidth]{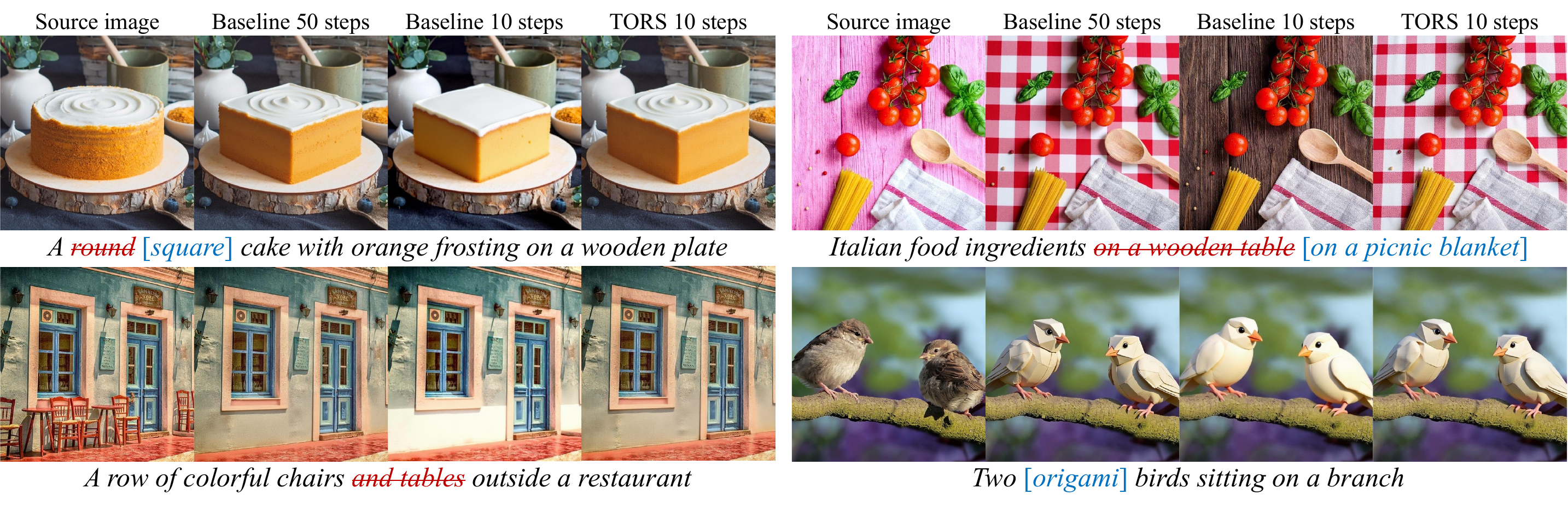}
    \caption{Qualitative results of image editing using Flux.1-Kontext.}
    \label{fig:quantitative_editing}
\end{figure*}

\textbf{Transferability}. We further demonstrate the transferability of our method by deploying \ourName schedules to downstream applications like image editing. Specifically, we \textbf{seamlessly reuse} the \ourName schedules computed for Flux on Flux.1-Kontext~\cite{labs2025flux}, one of the state-of-the-art image editing models. Quantitative evaluations on PIE-Bench~\cite{ju2023direct} (\Cref{tab:editing}) reveal that our method remains highly effective for image manipulation, significantly outperforming the default time schedule. As illustrated in \Cref{fig:quantitative_editing}, \ourName preserves superior layout consistency relative to 50-step reference generations.

\renewcommand{\arraystretch}{0.9}
\begin{table}[t]
    \caption{Performance of image editing evaluated on PIE-Bench using Flux.1-Kontext.}
    \label{tab:editing}
    \centering
    \fontsize{8}{10}\selectfont
    \begin{tabular}{l|c|cccc|cc}
        \toprule
        \multirow{2}{*}{Method} & Structure & \multicolumn{4}{c}{Background Preservation} & \multicolumn{2}{c}{CLIP Similarity} \\ 
         \cmidrule(lr){2-2} \cmidrule(lr){3-6} \cmidrule(lr){7-8}
        & Dist.$_{\times 10^3}$$\downarrow$ & PSNR$\uparrow$ & LPIPS$_{\times 10^3}$$\downarrow$ & MES$_{\times 10^4}$$\downarrow$ & SSIM$_{\times 10^2}$$\uparrow$ & Whole$\uparrow$ & Edit$\uparrow$ \\ 
        \midrule
        \rowcolor[gray]{0.9} Baseline $N=50$   & 47.52 & 28.33 & 59.32 & 90.17 & 91.67 & 25.71 & 22.65 \\
        Baseline $N=10$                        & 65.29 & 25.05 & 75.99 & 112.10 & 87.69 & 25.64 & 22.59 \\
        \textbf{TORS} $N=10$                         & \textbf{52.46} & \textbf{27.59} & \textbf{64.87} & \textbf{90.90} & \textbf{90.37} & \textbf{25.76} & \textbf{22.70} \\
        \bottomrule
    \end{tabular}
\end{table}


\begin{figure*}[t]
    \centering
    \begin{subfigure}[b]{0.495\textwidth}
        \includegraphics[width=\textwidth]{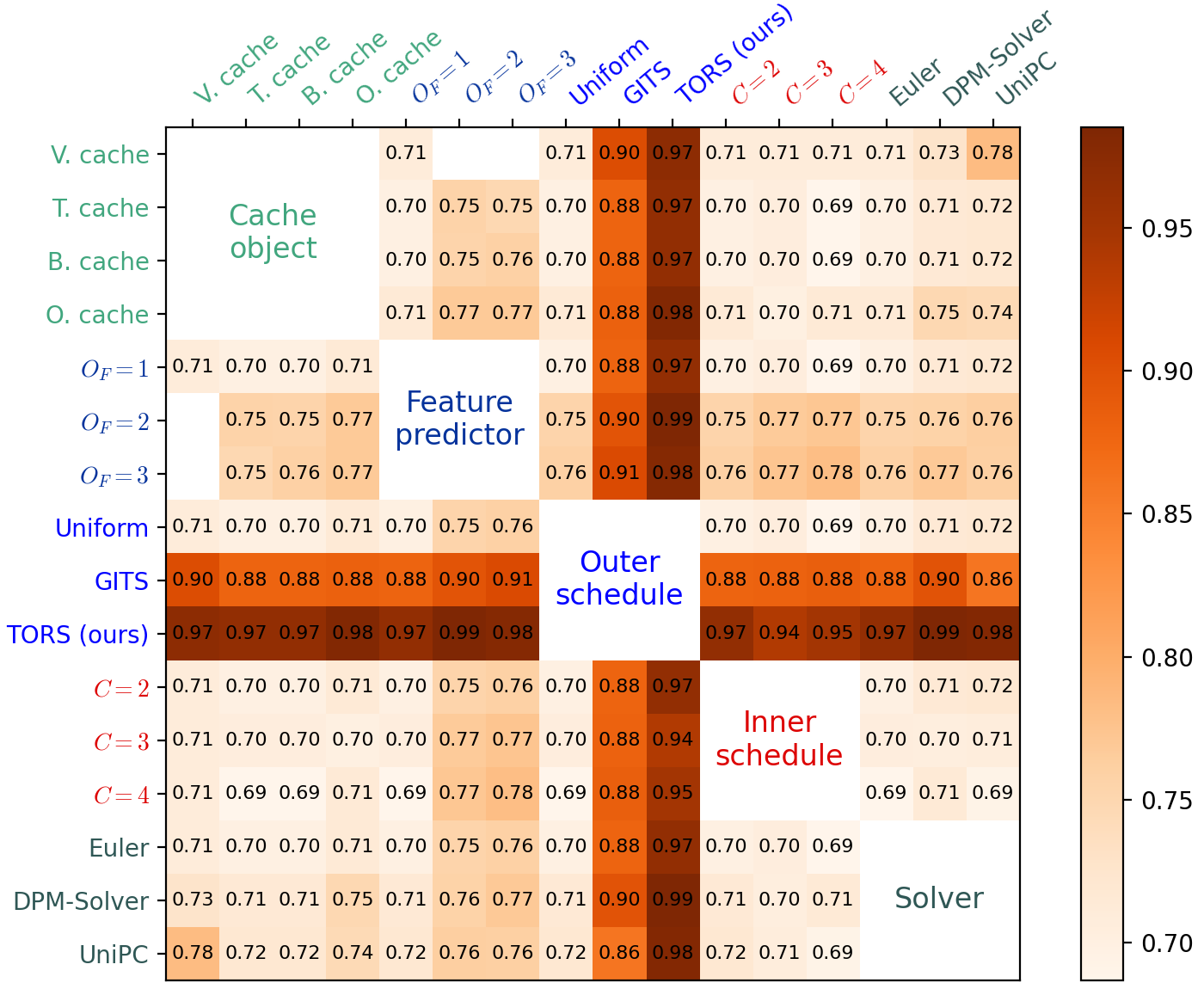}
        \caption{Compatibility evaluation on Flux.1-Dev.}
        \label{fig:compatibility_flux}
    \end{subfigure}
    \hfill
    \begin{subfigure}[b]{0.495\textwidth}
        \includegraphics[width=\textwidth]{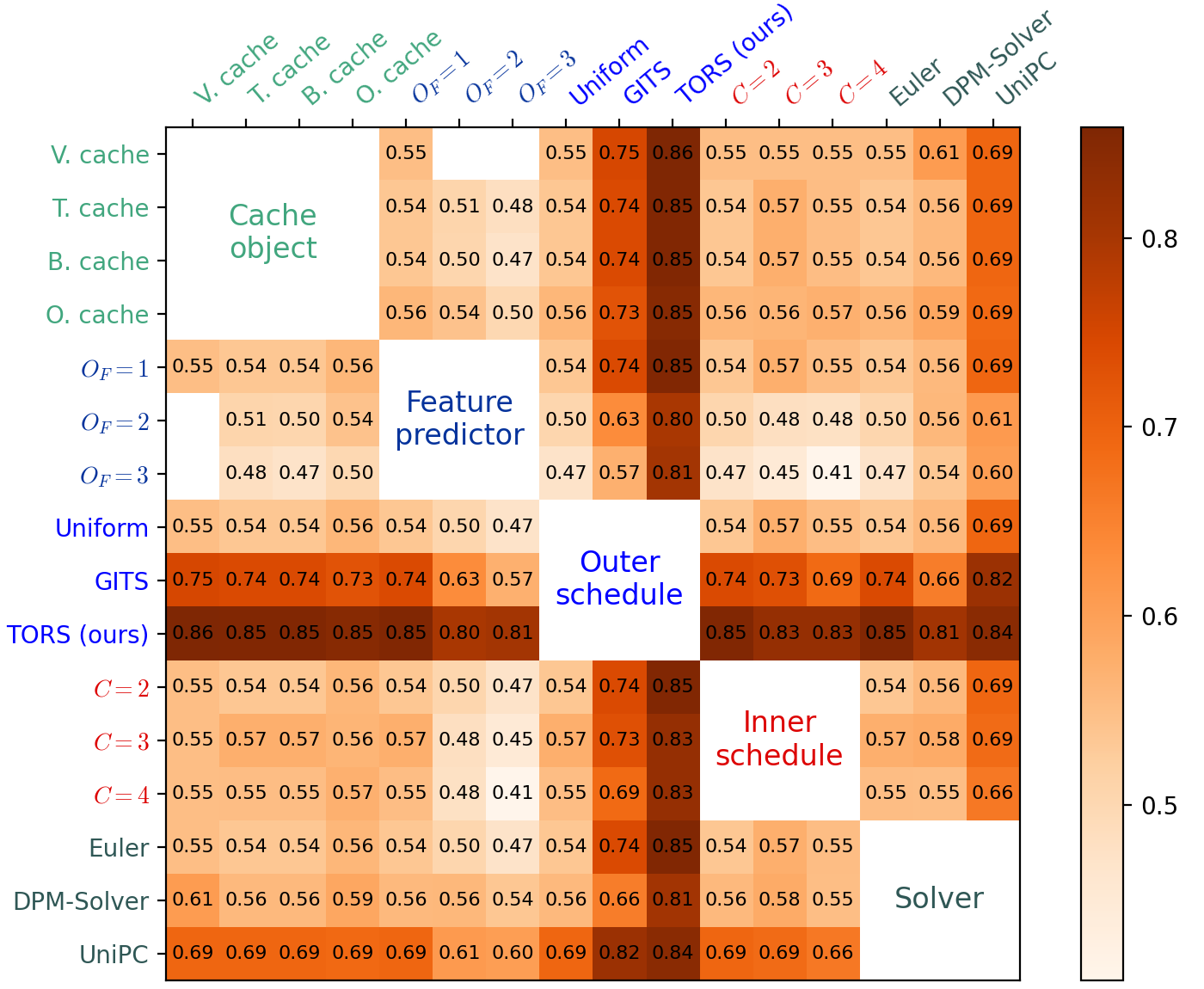}
        \caption{Compatibility evaluation on SD 3.5 medium.}
        \label{fig:compatibility_sd3}
    \end{subfigure}
    \caption{Compatibility evaluation on acceleration methods using Image Reward. 
    The performance can be enhanced by integrating appropriate acceleration strategies.}
    \label{fig:compatibility}
\end{figure*}

\subsection{Compatibility Evaluation}
\label{subsec:compatibility}
Beyond the univariate experiments presented in \Cref{sec:analyzing}, we conduct a thorough pairwise compatibility evaluation among the aforementioned sampling acceleration methods. The heatmap in \Cref{fig:compatibility} presents results under a sampling budget of 10 compute steps, covering all possible combinations of sampling acceleration groups. Several key observations emerge: 
(a) Our proposed \ourName demonstrates strong robustness, consistently achieving enhanced performance when combined with diverse acceleration strategies.
(b) Among all the acceleration strategies, the outer schedule predominantly drives the performance improvements.
(c) In most cases, fast solvers are compatible with other strategies and provide moderate performance gains.
(d) Neither the cache object nor the cache cycle significantly affects performance.
(e) The compatibility of the feature predictor with other strategies remains consistent within a single text-to-image model, though it may consistently enhance or degrade performance.

%% file: sec/conclusion.tex
\section{Conclusion}
Training-free sampling acceleration methods have improved the efficiency of text-to-image generation but have largely been developed in isolation. In this paper, we bridge this gap by elucidating and analyzing the design space of these methods from a unified perspective, through which we identify that the outer time schedule plays a dominant role in performance gains. We reveal the defects of the commonly used uniform schedule and introduce \ourName, an improved, efficient outer schedule that leverages the geometric properties of diffusion sampling trajectories. Extensive results demonstrate the superiority of \ourName. 
Furthermore, we examine the compatibility among the discussed acceleration methods, emphasizing the strong robustness of \ourName when integrated with various acceleration strategies.
Our work opens new avenues for sampling acceleration in state-of-the-art text-to-image diffusion models, and we believe the pursuit of ultimate training-free sampling acceleration through incorporation and refinement of multiple components is an important direction for future exploration.


%% file: sec/X_suppl.tex
\clearpage
\appendix

\title{Appendix}

\section{Related Works}

Text-to-image diffusion models have achieved impressive generative capabilities. However, their multi-step sampling process and the growing parameter scale largely intensify the resource requirements.
To address this issue, training-free sampling acceleration methods have successfully improved sampling efficiency through fast ODE solvers, efficient time schedules and feature caching.

\textbf{Fast solvers}.
By interpreting the sampling process of diffusion models through the PF-ODE framework~\cite{song2021sde,karras2022edm}, a range of fast solvers has been developed, leveraging classical numerical analysis methods. DDIM~\cite{song2021ddim} employs a first-order Euler discretization through a class of non-Markovian diffusion processes. EDM~\cite{karras2022edm} adopts Heun's second-order method. PNDM~\cite{liu2022pseudo} applies the linear multi-step method to solve the PF-ODE, integrating Runge-Kutta algorithms for the initial warm-up phase. Utilizing the semi-linear structure of the PF-ODE, DPM-Solver~\cite{lu2022dpm,lu2022dpmpp} and DEIS~\cite{zhang2023deis} are proposed by approximating the analytic solution using Taylor expansion and polynomial extrapolation, respectively. Additionally, UniPC~\cite{zhao2023unipc} introduces a unified predictor-corrector solver.

\textbf{Efficient time schedules}.
Training-free efficient time schedules typically involve manually designed heuristic schedules, such as the uniform log-SNR schedule~\cite{lu2022dpm} and the polynomial time schedules~\cite{karras2022edm}. 
Beyond these heuristic designs, Xue \etal~\cite{xue2024accelerating} propose an optimized time schedule by minimizing the discrepancy between the ground-truth solution of the ODE and an approximated solution obtained by ODE solvers. 
GITS~\cite{chen2024trajectory} introduces a geometry-inspired scheduling strategy that approximates the global truncation error by accumulating local errors. 
Additionally, there are works introducing additional training to optimize time schedules.
Watson \etal.~\cite{watson2021learning} optimizes time schedules utilizing the decomposable nature of the ELBO objective in diffusion models.
Wang \etal~\cite{wang2023learning} employ reinforcement learning to search for an efficient sampling schedule.
AutoDiffusion~\cite{li2023autodiffusion} employs evolutionary algorithms to explore the space of schedules to achieve an optimized FID score~\cite{heusel2017gans}.
AYS~\cite{sabour2024align} minimizes an upper bound on the Kullback-Leibler divergence to refine time scheduling.
Despite their effectiveness in low-resolution scenarios, the practicality and efficacy of existing time scheduling methods have not been validated in state-of-the-art text-to-image diffusion models.

\textbf{Feature caching} methods capitalize on the strong similarity of intermediate network features between adjacent sampling steps in diffusion models, dividing the sampling process into a series of ``compute-reuse'' cycles. During compute steps, the features are stored and subsequently utilized in reuse steps to bypass substantial computations in diffusion models. DeepCache~\cite{ma2024deepcache} caches the intermediate down-sample and up-sample block pairs in U-Net-based diffusion models. Li \etal~\cite{li2023faster} introduce an encoder propagation strategy. FORA~\cite{selvaraju2024fora} and $\Delta$-DiT~\cite{chen2024delta} extend feature caching to transformer-based diffusion models. TaylorSeers~\cite{liu2025reusing} employs Taylor expansion to predict the current feature using historical features. Additionally, there are training-based caching methods. Wimbauer \etal~\cite{wimbauer2024cache} propose an inner schedule for U-Net-based diffusion models through the relative absolute change along the sampling trajectory and adopt a scale-shift optimization to accommodate an aggressive caching ratio. Ma \etal~\cite{ma2024learning} further develop an optimized inner schedule through supervised learning.

\section{Experimental Details}

\subsection{Collection of Sampling Trajectories}
\label{subsec:app_collection}

\begin{figure}[t]
  \centering
  \begin{minipage}[b]{0.49\textwidth}
    \centering
    \includegraphics[width=\linewidth]{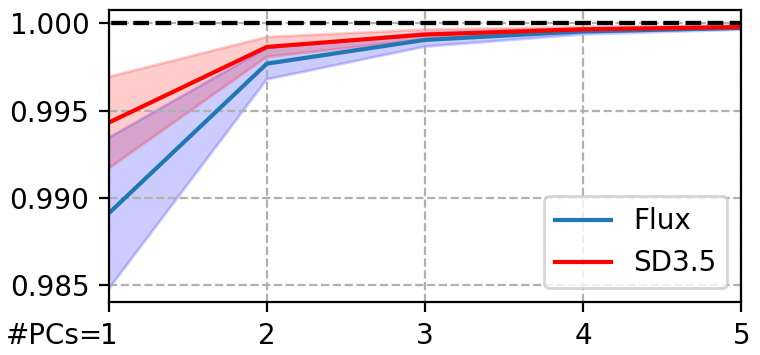}
    \caption{Explained variance by top principal components. The explained variance exceeds 99.9$\%$ with only three principal components.}
    \label{fig:explain}
  \end{minipage}
  \hfill
  \begin{minipage}[b]{0.49\textwidth}
    \centering
    \includegraphics[width=\linewidth]{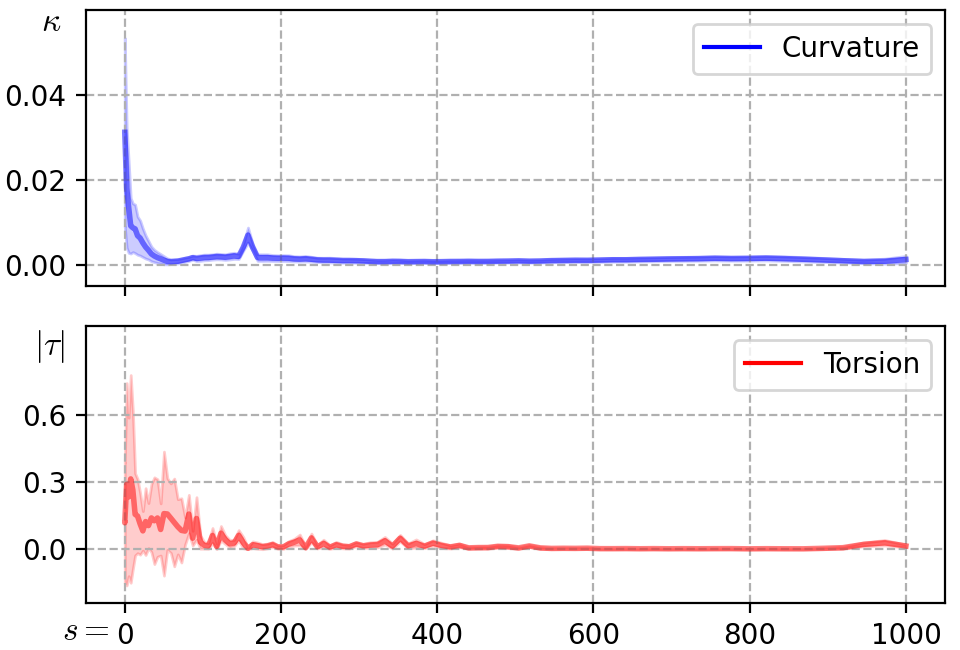}
    \caption{Calculated curvature and torsion statistics on Flux.1-Dev.}
    \label{fig:curv_tors_flux}
  \end{minipage}
\end{figure}

To demonstrate the geometric regularity of diffusion models and calculate their geometric properties, specifically curvature and torsion, we collect 100 sampling trajectories, each comprising 100 sampling steps, using text prompts randomly sampled from the MS-COCO dataset~\cite{lin2014microsoft}. We apply standard Principal Component Analysis (PCA) to the generated sampling trajectories and project them into three-dimensional space, where they exhibit strong regularity. The variance explained by the top principal components (PCs) is reported as the ratio of the sum of the squared top eigenvalues to the sum of all squared eigenvalues. As shown in \Cref{fig:explain}, the explained variance exceeds 99.9$\%$ with only three principal components. The total latency for trajectory collection is 40 minutes for Flux.1-Dev and 15 minutes for Stable Diffusion 3.5 medium, utilizing 4 NVIDIA RTX A6000 GPUs.

\begin{algorithm}[t]
\caption{Generate \ourName Schedule}
\label{alg:tors_generation}
\begin{algorithmic}[1]
\REQUIRE Diffusion model $\mathcal{M}$, number of collection trajectories $M$, collection steps $T$, target sampling steps $N$.

\STATE \textbf{Step 1: Trajectory Collection and Geometric Profiling}
\STATE Sample $M$ trajectories $\{\bfx_m\}_{m=1}^M$ from $\mathcal{M}$ using $T$ uniform steps.
\FOR{each trajectory $m = 1, \dots, M$}
    \STATE Project $\bfx_m$ into 3D coordinates $\mathbf{r}_m(t) = (x, y, z)$ using PCA.
    \STATE Compute arc-length $s$ via cumulative lengths: $s(t) = \int_0^t \|\dot{\mathbf{r}}_m(u)\| du$.
    \STATE Calculate curvature $\kappa_m(s)$ and torsion $\tau_m(s)$.
\ENDFOR

\STATE \textbf{Step 2: Statistic Aggregation}
\STATE Compute mean curvature $\bar{\kappa}(s)$ and mean absolute torsion $\bar{\tau}(s)$ across trajectories.
\STATE Compute the instantaneous rotation rate: $|\boldsymbol{\omega}(s)| = \sqrt{\bar{\kappa}(s)^2 + \bar{\tau}(s)^2}$.
\STATE Establish the mapping $t = f(s)$ between timestamp and arc-length.

\STATE \textbf{Step 3: Schedule Generation via Equidistribution}
\STATE Compute the total rotation $\Theta_{[0,S]} = \int_0^S |\boldsymbol{\omega}(s)| ds$.
\STATE Determine partitioning arc-lengths $\{s_i\}_{i=0}^N$ such that:
$$\int_0^{s_i} |\boldsymbol{\omega}(s)| ds = \frac{i}{N} \Theta_{[0,S]}, \quad \text{for } i = 0, \dots, N.$$
\STATE Map the dividing arc-lengths back to timestamps: $t_i = f(s_i)$.
\RETURN TORS time schedule $\mathcal{T} = \{t_0, t_1, \dots, t_N\}$.
\end{algorithmic}
\end{algorithm}

\subsection{Calculation of Curvature and Torsion}
\label{subsec:app_calculation}
When projecting the sampling trajectories into three-dimensional space using the top three PCs, we observe significant discrepancies in the magnitudes along different axes. These variations result in numerical oscillations when calculating the torsion $\tau=-\bfN\cdot \frac{\rmd \bfB}{\rmd s}$ using the naive forward difference method. To mitigate this issue, we implement an advanced method proposed by Lewiner \etal~\cite{LEWINER2005641}, which employs weighted least-squares fitting and local arc length approximation to enhance the accuracy. Given that the total arc length shows small variation among different projected sampling trajectories, we normalize all trajectories to a consistent total arc length of $S=1000$. In \Cref{fig:curv_tors_flux}, we present the statistics calculated on Flux.1-Dev which are averaged across 100 sampling trajectories.

\subsection{Implementation of Comparison Methods}
\label{subsec:app_implementation}
In \Cref{subsec:exp_ours}, we compare our proposed \ourName with cutting-edge training-free sampling acceleration methods. Below, we provide the implementation details.

For feature caching methods, we include FORA~\cite{selvaraju2024fora} and TaylorSeers~\cite{liu2025reusing}, both of which are representative methods tailored for transformer-based diffusion models. A total of 10 compute steps is ensured by setting $N=20$ and $C=2$. Both methods utilize Euler discretization, a uniform outer schedule and an operation cache in the original papers. Building upon these settings, FORA is recovered by setting $O_F=1$. For TaylorSeers, we use $O_F=2$. It is important to note that TaylorSeers employs the warm-up strategy described in \Cref{sec:analyzing} by default, which we identify as the critical factor improving performance in text-to-image generation. To accurately assess the effectiveness of the outer schedule, we disable this strategy in the comparison. While the proposed feature predictor yields only moderate performance gain for text-to-image generation, its effectiveness in class-conditional generation is noticeable, as observed in our preliminary experiments.

For solvers, we implement DPM-Solver~\cite{lu2022dpm} and UniPC~\cite{zhao2023unipc} using their default settings of $O_S=2$. Additionally, we apply a velocity cache and a uniform outer schedule for both of them.  

For the outer schedule, we compare our scheduling strategy with GITS~\cite{chen2024trajectory}, an advanced time scheduling method developed using dynamic programming. Regarding the hyperparameters for GITS, we set the teacher sampling steps to 100, the warm-up sample size to 64, and the DP coefficient to 0.9, which we find to deliver good results.

\section{Additional Discussions and Results}

\subsection{Further Discussion on Cache Object}
\label{subsec:app_cache_obj}

Implementing deeper caching may enhance performance, but it could also result in increased latency and VRAM demands. In \Cref{tab:latency_vram}, we present statistical data to examine the impact of different cache objects. All statistics are measured using a single NVIDIA RTX A6000 GPU. 


\renewcommand{\arraystretch}{0.9}
\begin{table}[t]
    \centering
    \caption{Latency and VRAM requirements on Flux.1-Dev with a batch size of 1. All statistics are measured on a single A6000 GPU. Velocity cache requires 38.88 GB, 41.44 GB and 44.00 GB of VRAM for batch sizes of 2, 3 and 4, respectively.}
    \label{tab:latency_vram}
    \fontsize{8}{10}\selectfont
    \begin{tabular}{l *{2}{w{r}{1cm} @{ / } w{r}{0.8cm} @{ / } w{r}{0.8cm}}}
        \toprule
        \multirow{2}{*}{Cache Object} & \multicolumn{3}{c}{Latency (s) $\downarrow$} & \multicolumn{3}{c}{VRAM (GB) $\downarrow$} \\
        \cmidrule(lr){2-4} \cmidrule(lr){5-7}
        & \multicolumn{3}{c}{$O_F = 1/2/3$} & \multicolumn{3}{c}{$O_F = 1/2/3$} \\
        \midrule
        V. cache & 9.85  & --    & --    & 36.32 & --    & --    \\
        T. cache & 9.90  & 9.92  & 9.94  & 36.38 & 36.44 & 38.61 \\
        B. cache & 10.03 & 10.41 & 10.64 & 37.94 & 39.56 & 41.17 \\
        O. cache & 10.47 & 10.87 & 11.19 & 38.48 & 40.63 & 42.78 \\
        \bottomrule
    \end{tabular}
\end{table}

\begin{figure}[h]
  \centering
  \begin{minipage}[b]{0.45\textwidth}
    \centering
    \includegraphics[width=\linewidth]{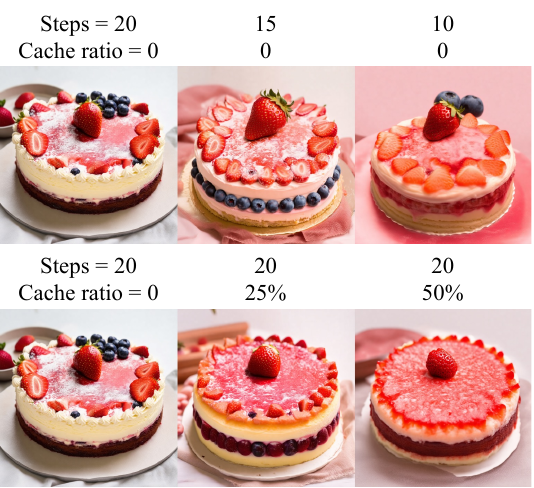}
    \caption{The effect of block-wise inner schedule~\cite{wimbauer2024cache} on SD3.5 medium. Cache ratio refers to the percentage of skipped transformer blocks. Prompt: \textit{A cake with strawberry and blueberry}.}
    \label{fig:bc_sd3_qualitative}
  \end{minipage}
  \hfill
  \begin{minipage}[b]{0.53\textwidth}
    \centering
    \includegraphics[width=\linewidth]{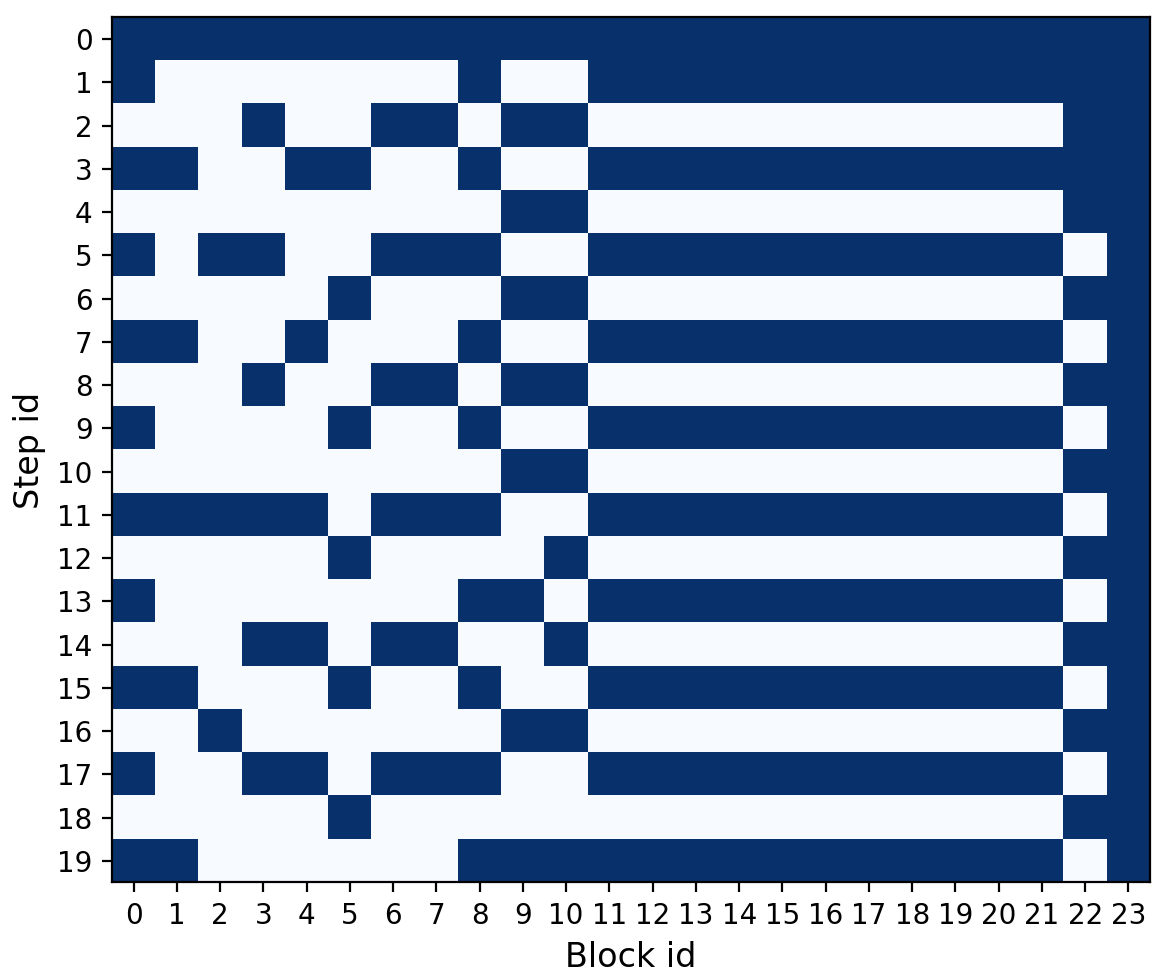}
    \caption{Block-wise inner schedule with cache ratio of 50$\%$. White blocks indicate skipped transformer blocks.}
    \label{fig:bc_sd3}
  \end{minipage}
\end{figure}

\begin{figure*}[h]
    \centering
    \begin{subfigure}[b]{0.495\textwidth}
        \includegraphics[width=\textwidth]{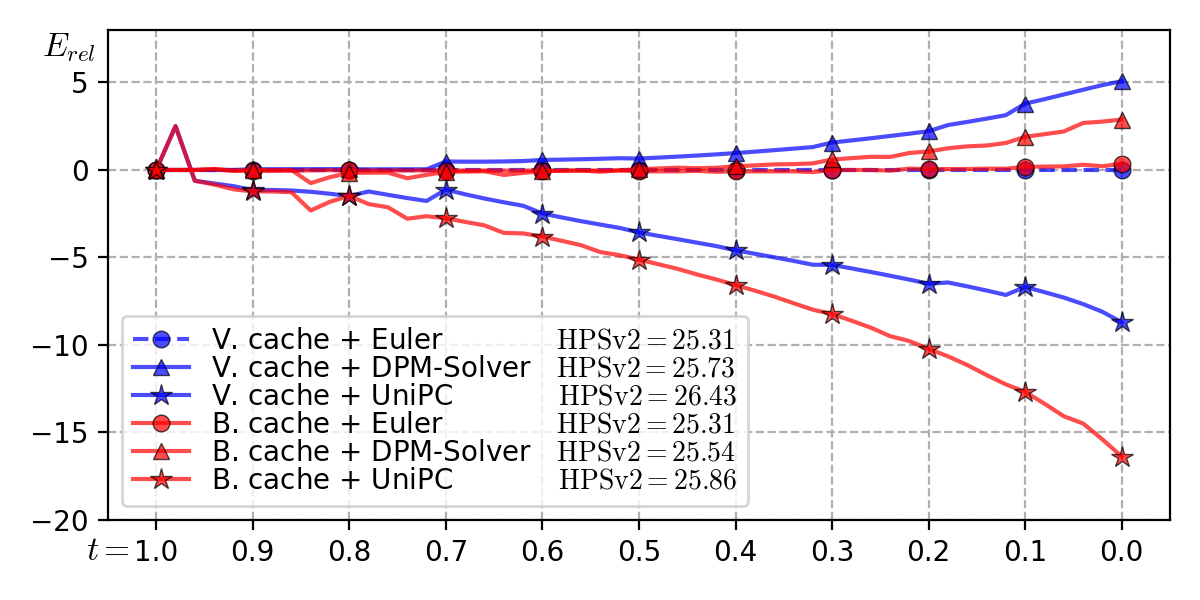}
        \caption{Solver.}
        \label{fig:compare_sd3_solver}
    \end{subfigure}
    \hfill
    \begin{subfigure}[b]{0.495\textwidth}
        \includegraphics[width=\textwidth]{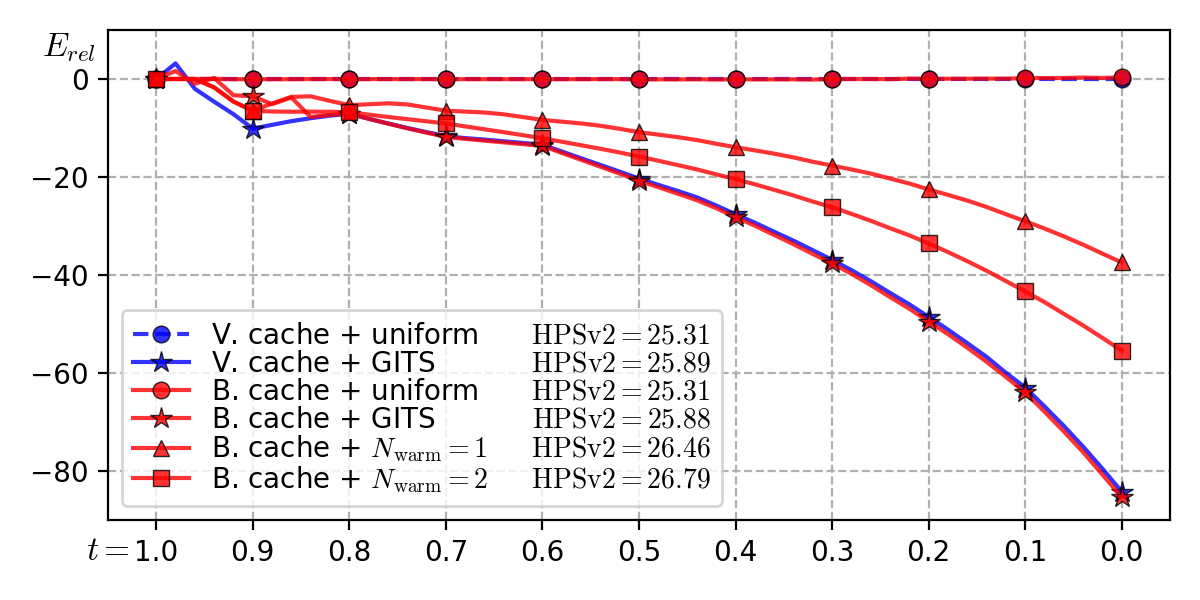}
        \caption{Outer schedule.}
        \label{fig:compare_sd3_outer}
    \end{subfigure}
    \hfill
    \begin{subfigure}[b]{0.495\textwidth}
        \includegraphics[width=\textwidth]{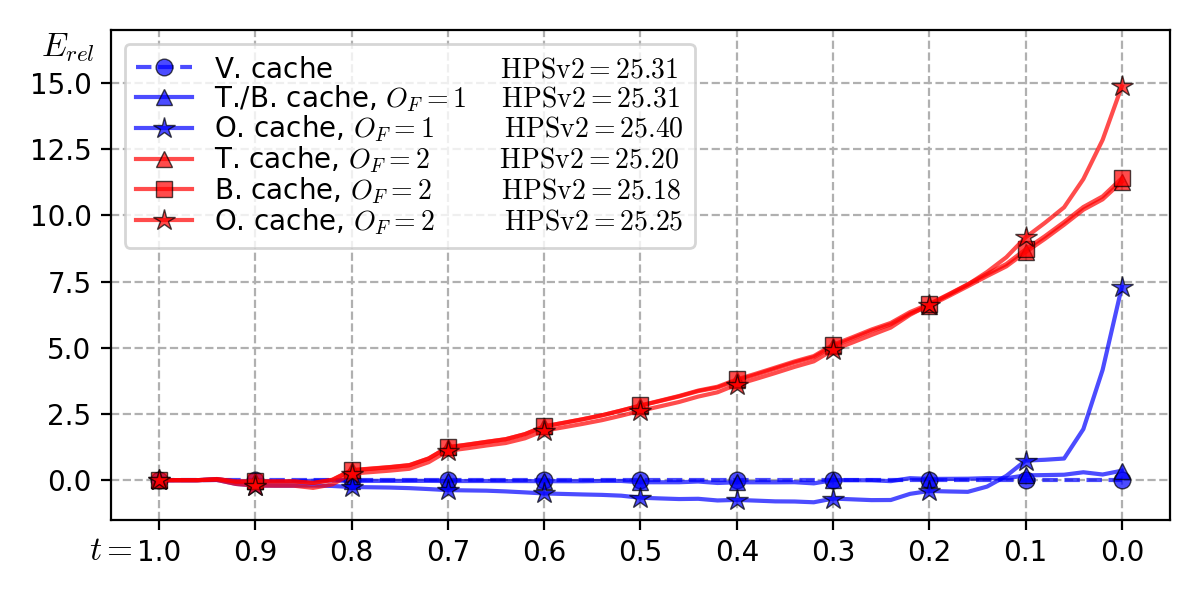}
        \caption{Cache object.}
        \label{fig:compare_sd3_object}
    \end{subfigure}
    \hfill
    \begin{subfigure}[b]{0.495\textwidth}
        \includegraphics[width=\textwidth]{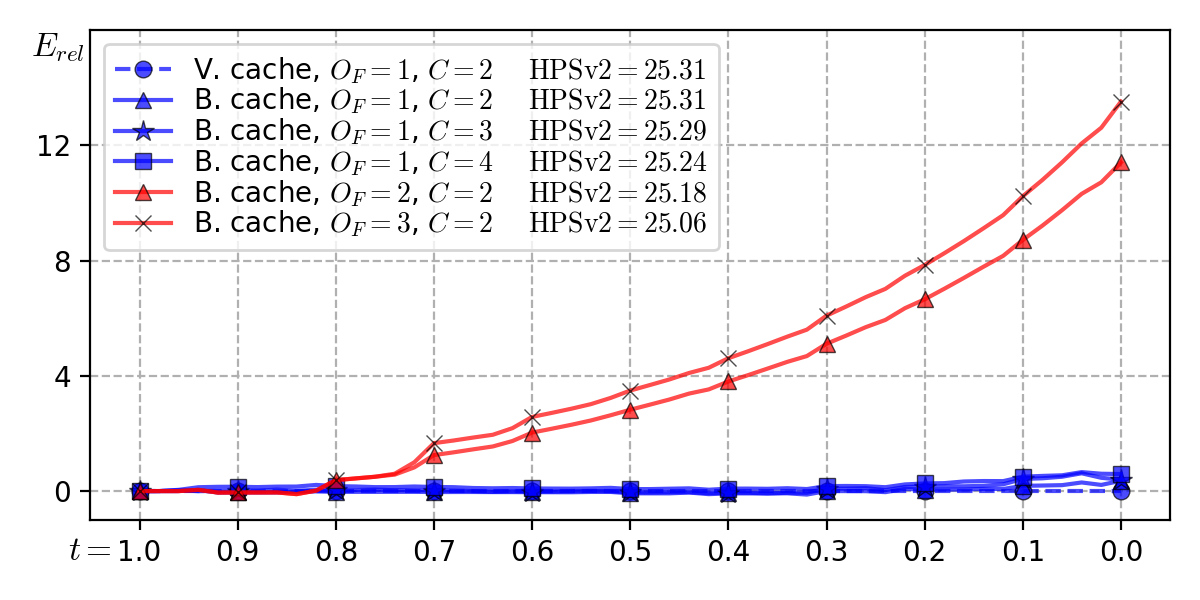}
        \caption{Inner schedule and feature predictor.}
        \label{fig:compare_sd3_pred_inner}
    \end{subfigure}
    \caption{Comprehensive experiments examining the impact of each acceleration method on sampling acceleration. The experiments are conducted on Stable Diffusion 3.5 medium model~\cite{esser2024scaling}. The outer schedule is identified as the most influential factor.}
    \label{fig:compare_sd3}
\end{figure*}

\subsection{Further Discussion on Inner Schedule}
\label{subsec:app_inner}
Wimbauer \etal~\cite{wimbauer2024cache} propose a block-wise scheduling strategy that assigns a distinct inner schedule for different U-Net blocks. The strategy is derived by analyzing the relative absolute changes along the sampling trajectories. We reimplement this method in transformer-based diffusion models but observe degraded results, as shown in \Cref{fig:bc_sd3_qualitative}. We hypothesize that this degradation stems from the errors accumulated at each sampling step, since only the first step completes accurate model inference (see \Cref{fig:bc_sd3}).



\subsection{Additional Quantitative Results}
\label{subsec:app_quantitiative}
In this section, we replicate the experiments conducted with Flux.1-Dev~\cite{flux} in \Cref{sec:analyzing} and apply them to Stable Diffusion 3.5 medium~\cite{esser2024scaling}. We include a thorough examination of each component of sampling acceleration methods, as detailed in \Cref{fig:compare_sd3}. Consistent with the findings from the Flux model, efficient solvers and outer schedules contribute significantly to performance improvements. However, in the case of Stable Diffusion 3.5, the application of feature caching methods does not yield performance gains. This discrepancy highlights the model-specific effectiveness of feature caching methods and underscores the necessity for tailored optimization strategies depending on the text-to-image model in use, which we leave to future work.

\begin{figure*}[t]
    \centering
    \includegraphics[width=\textwidth]{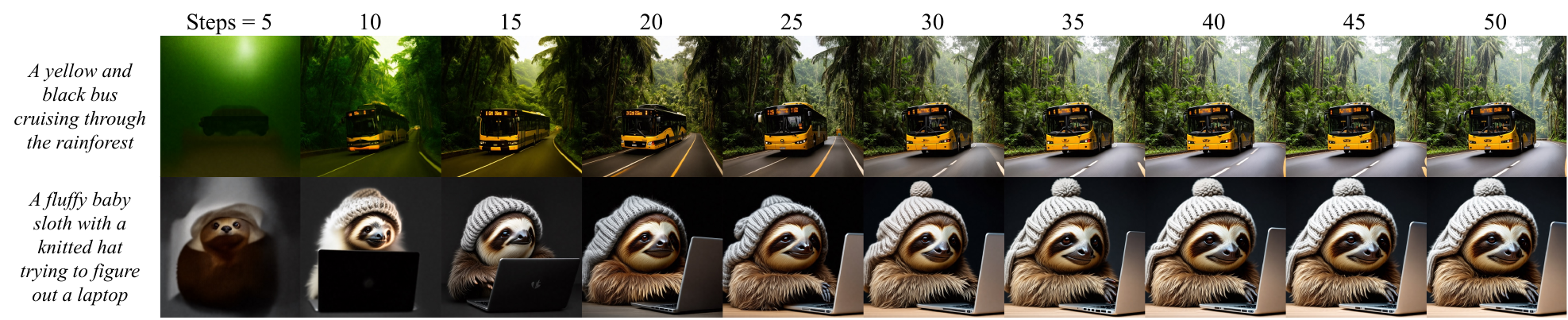}
    \caption{Text-to-image generation with Stable Diffusion 3.5 medium~\cite{esser2024scaling} using the default uniform outer schedule.}
    \label{fig:flawed_sd3}
\end{figure*}

\begin{figure*}[t]
    \centering
    \begin{subfigure}[b]{0.33\textwidth}
        \includegraphics[width=\textwidth]{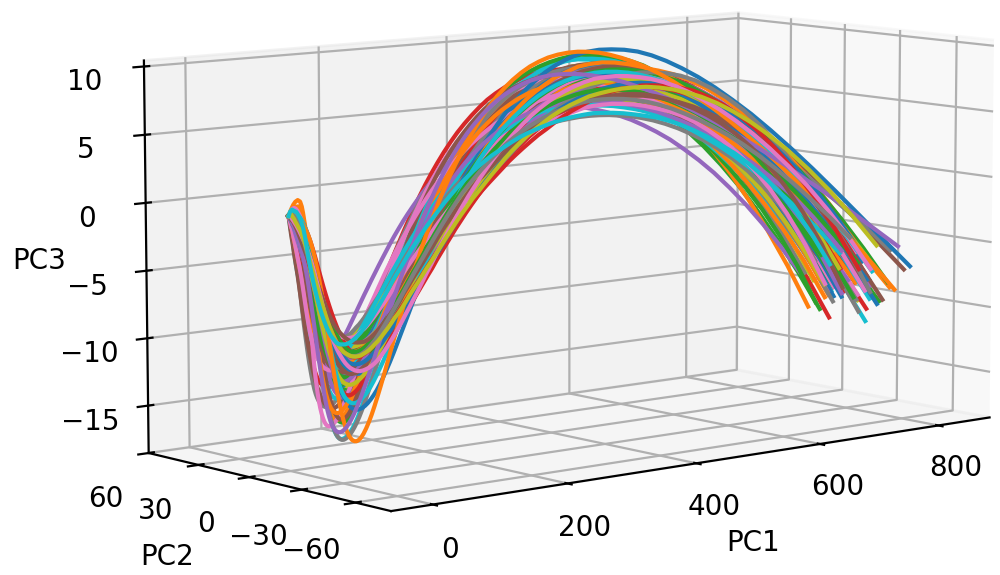}
        \caption{Trajectory regularity on Stable Diffusion 3.5.}
        \label{fig:regularity_sd3}
    \end{subfigure}
    \hfill
    \begin{subfigure}[b]{0.29\textwidth}
        \includegraphics[width=\textwidth]{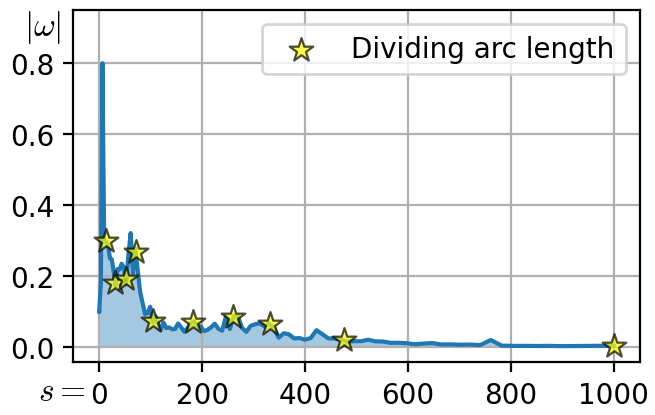}
        \caption{Selected dividing arc lengths for \ourName.}
        \label{fig:selected_sd3}
    \end{subfigure}
    \hfill
    \begin{subfigure}[b]{0.35\textwidth}
        \includegraphics[width=\textwidth]{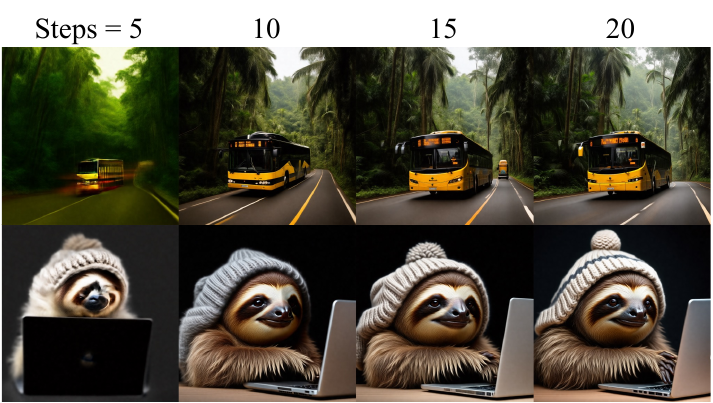}
        \caption{\ourName yields faster structural convergence.}
        \label{fig:ours_qualitative_sd3}
    \end{subfigure}
    \caption{Trajectory regularity and the effectiveness of our proposed \ourName on Stable Diffusion 3.5 medium.}
    \label{fig:regularity_app}
\end{figure*}

Similar to the Flux model, in \Cref{fig:flawed_sd3}, we observe a slow structural convergence for Stable Diffusion 3.5, which again highlights the deficiencies of the default uniform outer schedule. We project the sampling trajectories generated by Stable Diffusion 3.5 and visualize them in \Cref{fig:regularity_sd3}, where a strong geometric regularity is evident. Following the procedure described in \Cref{subsec:ours}, we calculate curvature and torsion of the projected sampling trajectories. By selecting dividing arc lengths that ensure a constant change in the total rotation, we obtain our scheduling strategy \ourName by mapping these arc lengths back to timestamps. A 10-step example is shown in \Cref{fig:selected_sd3}. In comparison to the flawed uniform outer schedule, we provide qualitative results of our \ourName with Stable Diffusion 3.5, using identical prompts and starting Gaussian noises as in \Cref{fig:flawed_sd3}. The results displayed in \Cref{fig:ours_qualitative_sd3} demonstrate that \ourName significantly improves performance and achieves faster structural convergence.



\subsection{Prompts}
\label{subsec:app_prompts}
We present the text prompts used in this paper.

Text prompts used in \Cref{fig:teaser}:
{\itshape
\begin{itemize}
    \item Photo of an athlete cat explaining it's latest scandal at a press conference to journalists.
    \item A painting by Grant Wood of an astronaut couple, American gothic style.
    \item A banana on the left of an apple.
    \item A sign that says ``Text to Image''.
    \item A laptop on top of a teddy bear.
    \item A couple of glasses are sitting on a table.
    \item A mechanical or electrical device for measuring time.
    \item Hovering cow abducting aliens.
\end{itemize}
}

Text prompts used in \Cref{fig:qualitative_flux}:
{\itshape
\begin{itemize}
    \item A tiger in a lab coat with a 1980s Miami vibe, turning a well oiled science content machine, digital art.
    \item A pear cut into seven pieces arranged in a ring.
    \item A cat on the right of a tennis racket.
    \item An illustration of a large red elephant sitting on a small blue mouse.
    \item A large thick-skinned semiaquatic African mammal, with massive jaws and large tusks.
    \item A storefront with ``Deep Learning'' written on it.
\end{itemize}
}

Text prompts used in \Cref{fig:qualitative_sd3}:
{\itshape
\begin{itemize}
    \item New York Skyline with 'Diffusion' written with fireworks on the sky.
    \item A black colored dog.
    \item A panda making latte art.
    \item A maglev train going vertically downward in high speed, New York Times photojournalism.
    \item A car playing soccer, digital art.
    \item Hyper-realistic photo of an abandoned industrial site during a storm.
\end{itemize}
}

\subsection{Additional Qualitative Results}
\label{subsec:app_qualitative}
\begin{figure*}[t]
    \centering
    \includegraphics[width=\textwidth]{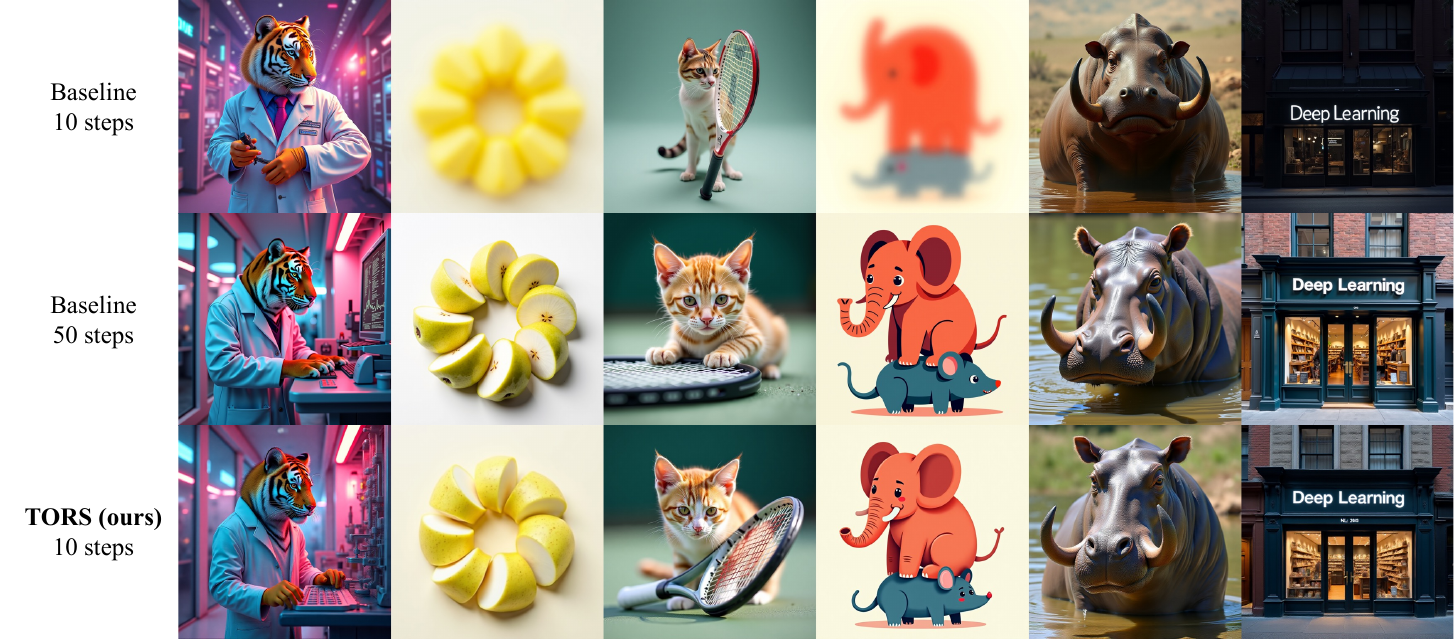}
    \caption{Additional qualitative results using Flux.1-Dev~\cite{flux}.}
    \label{fig:qualitative_flux}
\end{figure*}

\begin{figure*}[t]
    \centering
    \includegraphics[width=\textwidth]{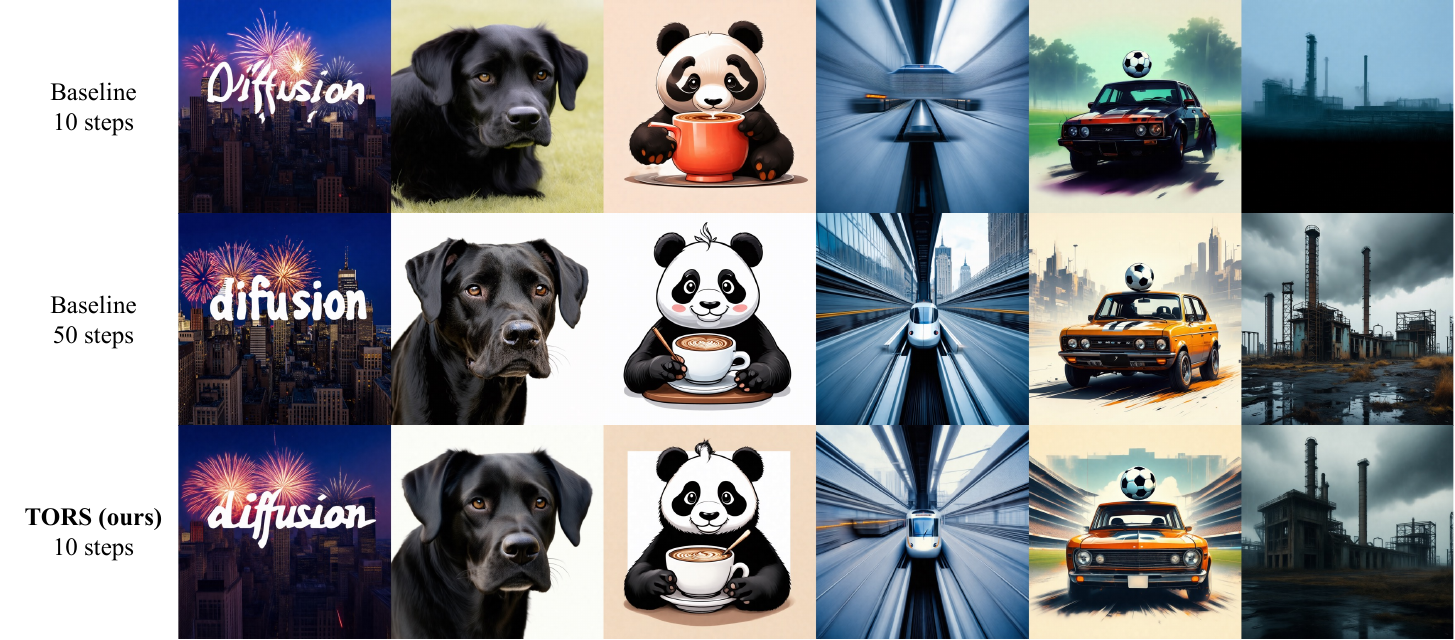}
    \caption{Additional qualitative results using Stable Diffusion 3.5 medium~\cite{esser2024scaling}.}
    \label{fig:qualitative_sd3}
\end{figure*}

We provide more qualitative results comparing the baseline uniform outer schedule with our proposed \ourName. The results for both Flux.1-Dev and Stable Diffusion 3.5 are shown in \Cref{fig:qualitative_flux} and \Cref{fig:qualitative_sd3}. Our proposed \ourName demonstrates significant improvements when operating under constrained sampling budgets.

\clearpage

